\documentclass{article}
\usepackage[left=1in,top=1in,right=1in,bottom=1in,letterpaper]{geometry}

\usepackage{authblk}

\usepackage[round]{natbib}

\usepackage{amsmath,amssymb,amsthm,mathtools}
\usepackage{graphicx,booktabs,algorithm,algorithmic,hyperref,url,natbib,multirow,multicol, enumitem}
\usepackage[table]{xcolor}

\hypersetup{
    colorlinks=true,
    linkcolor=black,
    citecolor=blue,
    urlcolor=blue
}

\newtheorem{theorem}{Theorem}
\newtheorem{assumption}{Assumption}
\newtheorem{proposition}{Proposition}
\newtheorem{lemma}{Lemma}

\newtheorem{definition}{Definition}

\usepackage{xspace}

\newcommand{\Eb}{\mathbb{E}}
\newcommand{\qp}{q_\phi}

\newcommand{\Dr}{\mathcal{D}_\rho}
\newcommand{\cvm}{\mathrm{CvM}}

\newcommand{\qy}{q_\phi(\cdot | y)}
\newcommand{\tcvm}{TARP-CvM\xspace}

\definecolor{figblue}{HTML}{2C5F8A}\definecolor{figred}{HTML}{C0392B}
\newcommand{\mkfilled}{\textcolor{figblue}{$\bullet$}}\newcommand{\mkhollow}{\textcolor{figblue}{$\circ$}}\newcommand{\mkred}{\textcolor{figred}{$\circ$}}

\definecolor{lightblue}{RGB}{225,240,250}

\definecolor{forest}  {rgb}{0,.5,0} 
\newif\iffinal
\finaltrue 
\iffinal
    \newcommand{\todo}[1]{}
    \newcommand{\becca}[1]{}
    \newcommand{\ben}[1]{}
    \newcommand{\cc}[1]{}
\else
    \newcommand{\todo}[1]{{\textcolor{red}{[to do: #1]}}}
    \newcommand{\becca}[1]{{\color{forest}#1}}
    \newcommand{\ben}[1]{{\color{magenta}Ben: #1}}
    \newcommand{\cc}[1]{{\color{orange}CC: #1}}
\fi

\usepackage[capitalize]{cleveref}
\Crefname{assumption}{Assumption}{Assumptions}
\Crefname{figure}{Fig.}{Figs.}
\Crefname{equation}{}{}
\creflabelformat{equation}{#2(#1)#3}
\renewcommand{\eqref}[1]{\cref{#1}}

\title{Training Calibrated Conditional Generative Models with a Sample-Only Coverage Loss}

\author[1,2,3]{Benjamin Remy}
\author[4]{Raphael Rossellini}
\author[1,2,3]{Chihway Chang}
\author[1,4,5,6]{Rebecca Willett}

\affil[1]{NSF-Simons AI Institute for the Sky (SkAI)}
\affil[2]{Department of Astronomy and Astrophysics, University of Chicago}
\affil[3]{Kavli Institute for Cosmological Physics, University of Chicago}
\affil[4]{Department of Statistics, University of Chicago}
\affil[5]{NSF-Simons National Institute for Theory and Mathematics in Biology (NITMB)}
\affil[6]{Department of Computer Science, University of Chicago}

\begin{document}

\maketitle

\begin{abstract}
  Conditional generative models have become the tool of choice for sampling posteriors in scientific inverse problems, where we seek to sample images or signals conditioned on a set of indirect and noisy measurements. Yet they are often overconfident, which is a critical failure in scientific applications where accurate uncertainty quantification is essential. Existing regularizers fall short in complementary ways: differentiable coverage objectives require the model's density and do not identify the posterior, while strictly proper scoring rules such as the energy score need only samples but do not directly enforce coverage. We close this gap by introducing a differentiable loss function based on the expected coverage probability, estimated from samples alone. It applies to any conditional generative model, with no need for tractable densities or architectural changes. We prove that it is a strict divergence, vanishing if and only if the model equals the true posterior. On simulation-based inference benchmarks, it matches a density-based coverage loss without evaluating the density. On high-dimensional imaging inverse problems, it yields better-calibrated posteriors than scoring-rule regularizers, with little change in the reconstruction quality of point estimates.
\end{abstract}

\section{Introduction}

Inverse problems are ubiquitous across scientific domains, including medical imaging \citep{liu2009calculation},  astrophysics \citep{adam2022posterior, remy2023probabilistic}, engineering \citep{melia2025multi}, and economics \citep{horowitz2014ill}. Measurements 
of the signal of interest
are typically acquired indirectly, only partially, and corrupted by noise, making the recovery of the underlying parameters ill-posed. Many different parameters can explain the same observations, so 
we seek to sample from the full range of admissible signals; such samples enable uncertainty quantification.

Bayesian inference provides a principled framework for this, modeling the set of all plausible solutions through the posterior distribution of the parameters given the observations. Deep generative models have made it possible to learn this posterior from pairs of parameters and corresponding measurements. Normalizing Flows \citep[NF,][]{pmlr-v37-rezende15, JMLR:v22:19-1028}, generative adversarial networks \citep[GANs,][]{goodfellow2014generative}, and more recently diffusion \citep{sohl2015deep, ho2020denoising, song2021scorebased} and flow \citep{lipman2022flow, albergo2025stochastic} models have all been successful at conditional sampling. However, each of these models has known failure modes that produce unreliable posteriors. GANs are prone to mode collapse \citep{metz2016unrolled}, and NFs have been shown to be frequently overconfident for posteriors \citep{hermans2022a}. Diffusion models, while achieving remarkable sample quality, have also been found to under-represent the uncertainty of the true posterior \citep{yao2026posteriorbench}.

We seek models with valid \emph{conditional} coverage: for every measurement $y$ and every level $\alpha$, whenever the model assigns probability $\alpha$ to a region of signals, the true signal should fall in that region with probability $\alpha$ under the posterior $p(\theta \mid y)$, that is, $\alpha \cdot 100\%$ of the time over hypothetical draws of the true signal conditioned on that measurement. Requiring conditional coverage for all regions is equivalent to the learned posterior estimate $q_\phi(\cdot \mid y)$ equaling the true posterior $p(\cdot \mid y)$, but it cannot be checked directly, since a dataset typically contains a single ground-truth signal per measurement.

Two families of training objectives address this problem, each with a complementary weakness. \emph{Coverage objectives} \citep{delaunoy2022towards, falkiewicz2023} add to the training loss a differentiable measure of the estimator's coverage, typically computed on its highest probability density (HPD) regions, or a constraint favoring conservative estimators. They target coverage directly, but they have two limitations. Computing HPD regions requires evaluating the estimator density, which restricts these methods to architectures with tractable likelihoods, such as NFs, and excludes diffusion models and GANs. Moreover, they enforce marginal coverage, which does not identify the posterior: many distributions, including the prior, have valid marginal coverage. \emph{Strictly proper scoring rules}, such as the continuous ranked probability score (CRPS) or the energy score, require only samples and are uniquely minimized at the posterior, but they do not target coverage, and in practice they can yield poorly calibrated models due to finite-sample effects, model misspecification and the geometry of the loss landscape (\cref{sec:related}).

The basic idea is to assess a conditional generative model by asking how much of its probability mass lies closer to a reference point than the true signal does. For a given measurement $y$, choose a reference point $\psi(y)$ and consider balls centered at this point. The model assigns a probability mass to each ball, and we can check how often the true signal falls inside it. If the model is correct, the two agree at every probability level, and we measure departures from this agreement with a Cramér--von Mises (CvM) discrepancy. This construction builds on TARP (Tests of Accuracy with Random Points, \citealp{lemos2023sampling}), and we call the resulting loss \tcvm.
The converse depends on how reference points are chosen. A reference rule assigns a reference point to every measurement, and a single rule is not enough, even one that depends on the measurement. Its coverage is averaged over measurements, so over-coverage at some measurements can balance under-coverage at others. We therefore compute the discrepancy separately for many rules and average the discrepancies, rather than pooling their coverage. Our main theoretical result shows that if the rules can place reference points anywhere and can differ across regions of measurement space, this average vanishes only when the model equals the true posterior.
Our contributions are:
\begin{itemize}
  \item \textbf{Theory:} a sufficient condition on the distribution of measurement-dependent reference rules under which our \tcvm loss vanishes if and only if the model equals the true posterior.
  \item \textbf{Algorithm:} a sample-only, differentiable estimator of \tcvm that applies to any conditional generative model, together with a practical construction of reference rules that satisfies this condition.
  \item \textbf{Experiments:} on simulation-based inference, \tcvm matches a density-based coverage loss without evaluating the density; on high-dimensional imaging inverse problems, it is the only regularizer we tested that brings the posterior into the Mira null band on all three problems, with little change in the quality of point estimates.
\end{itemize}

\paragraph{Problem formulation.}
We seek to recover parameters $\theta \in \Theta \subseteq \mathbb{R}^{d_\theta}$ from indirect measurements $y \in {\cal Y} \subseteq \mathbb{R}^{d_y}$. 
Knowledge of the data likelihood $p(y \mid \theta)$ may be used to generate training samples but is not otherwise required. There may exist a forward model $\mathcal{A}$ such that $y = \mathcal{A}(\theta) + \varepsilon$, where $\varepsilon$ is a noise vector, and together ${\cal A}$ and the distribution of $\varepsilon$ control the data likelihood $p(y\mid\theta)$. This setting encompasses a wide range of problems: in inverse problems, $\theta$ may be a high-dimensional image of a patient's anatomy, $y$ the raw measurements collected by an MRI scanner, and $\mathcal{A}$ a model of the scanner's measurement process; in simulation-based inference, $\theta$ may be the parameters of a physics simulation, with $\mathcal{A}$ the simulator itself, in which case we seek to infer $\theta$ from observations $y$.
We seek to sample from the posterior distribution $p(\theta \mid y)$, but cannot evaluate the prior density $p(\theta)$. We have only samples, a training set $\{(\theta_1, y_1), \ldots, (\theta_n, y_n)\}$ of $n$ pairs drawn from the joint distribution $p(\theta, y)$, so that the $\theta_i$ are samples from the prior. Our strategy is to train a conditional generative model, denoted $\qy$, which takes as input a measurement vector $y$ and a random seed and generates a sample $\theta$, where $\phi \in \mathbb{R}^p$ is the vector of generative model parameters (e.g., neural network weights) to be learned.

\section{Related work}\label{sec:related}

\begin{figure*}[h]
    \centering
    \includegraphics[width=1.\linewidth]{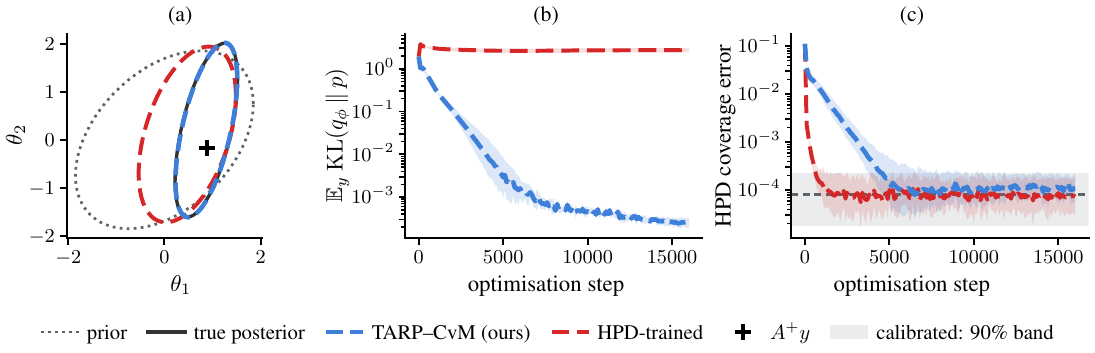}
    \caption{Gaussian linear inverse problem, $\theta\in\mathbb{R}^8$. (a) 2-d slice of $68\%$ level sets of the prior, the true posterior, and models trained with the HPD loss and with \tcvm alone. (b) Expected KL to the true posterior during training. (c) HPD coverage error during training: it cannot distinguish the two models, although their KL differs strongly. Setup and details in Section~\ref{sec:fig1} and Appendix~\ref{sec:appendix-toy}.}
    \label{fig:fig1}
\end{figure*}

\paragraph{Generative models for inverse problems}
\textit{Model-agnostic} methods train a conditional generative model on paired samples $(\theta_i, y_i)$, without using the forward model $\mathcal{A}$. \textit{Model-adaptive} methods instead combine a generative prior over $\theta$, pre-trained without measurements, with knowledge of $\mathcal{A}$ at sampling time, as in Diffusion Posterior Sampling \citep{chung2022diffusion}. We focus on the model-agnostic setting, but the core ideas translate naturally to model-adaptive samplers.

\paragraph{Coverage tests} 
Two standard approaches for empirically testing coverage of $q_\phi$ are Highest Posterior Density \citep[HPD,][]{hermans2022a} testing and Tests of Accuracy with Random Points \citep[TARP,][]{lemos2023sampling}. HPD testing constructs, for each test point $y$, the smallest region $C \subset \mathbb{R}^{d_\theta}$ under $q_\phi(\cdot \mid y)$ that contains $\alpha$ probability mass, and then checks whether the true $\theta$ falls inside that region at the expected rate. However, HPD requires evaluating the density of $q_\phi$, which is often unavailable for generative models, and valid HPD coverage is not a sufficient condition for accuracy: a model can pass the HPD test while still differing from the posterior. \cref{fig:fig1} illustrates the limitations of using HPD to train a conditional generative model vis-\`{a}-vis our proposed measure, \tcvm; details on this experiment are in \cref{sec:fig1}.

TARP instead constructs regions as balls centered at randomly chosen reference points, measuring coverage by the fraction of samples from $q_\phi(\cdot \mid y)$ that fall closer to the reference point than the true $\theta$ does. \citet{lemos2023sampling} show that correct TARP coverage for every assignment of reference points to measurements is necessary and sufficient for the estimator to equal the posterior, so TARP can detect errors that HPD-based tests miss, while requiring only samples from $\qp$ rather than density evaluations. The key distinction is that HPD regions are implicitly centered by the geometry of the estimated density, whereas TARP regions can be centered at arbitrary reference points. This flexibility lets TARP probe aspects of the posterior that HPD coverage cannot capture. Which assignments are needed in practice is left open; we make this precise in Section~\ref{sec:tarp-cvm}. The output of TARP is a \textit{coverage curve} that plots the empirical coverage rate against the nominal credibility level $\alpha$, with a perfectly calibrated model lying on the diagonal. Although correct TARP coverage for every possible assignment of reference points characterizes the posterior, this does not immediately provide a practical training objective. In particular, averaging coverage over randomly chosen reference points can conceal errors, because overcoverage and undercoverage may cancel. Our contribution is to identify a tractable family of reference rules whose individual coverage discrepancies can be averaged without losing the ability to identify the posterior.

Because TARP produces a coverage curve rather than a scalar, it does not directly support quantitative comparisons between models. Mira \citep{sharief2026mira} provides such a scalar. Over many randomly drawn regions and joint samples $(\theta, y) \sim p(\theta, y)$, it estimates the probability that the true $\theta$ falls in a region given the fraction of samples from $q_\phi(\cdot \mid y)$ that fall in it. Under calibration, the Mira score equals $2/3$, and values below this level indicate overconfidence. Since our loss is built on TARP, evaluating it with TARP coverage alone would be partly circular. We therefore use Mira as an independent metric: it relies on its own random regions, is not optimized by any of the methods we compare, and provides a calibrated reference value with a null band that accounts for sampling noise.

\paragraph{Energy Score and CRPS}
Using MSE as a loss function assesses a model's ability to correctly estimate the conditional mean, but it does not reward modeling the spread of possibilities correctly and can prefer overconfident generators. The Energy Score \citep[ES, ][]{gneiting2007strictly}
\begin{equation*}
    \mathrm{ES}(\qy, \theta) = 2 \mathbb E_{\hat \theta \sim \qy} \|\hat \theta- \theta\|_2 - \mathbb E_{\hat \theta, \hat \theta' \sim \qy} \|\hat \theta- \hat \theta'\|_2,
\end{equation*}
addresses this by penalizing samples far from the true $\theta$ while rewarding spread. It is strictly proper on distributions with finite first moment, with expectation under $p(\cdot\mid y)$ uniquely minimized at $\qp(\cdot\mid y)=p(\cdot\mid y)$, and it can be estimated from samples alone. Its univariate case is the continuous ranked probability score (CRPS). In high dimension, the CRPS is often summed over coordinates (pCRPS), which parallelizes across pixels, but only identifies the marginals of the posterior. 
Strict propriety says that the population objective has the true posterior as its unique global minimizer. It does not guarantee that training finds this minimizer. The loss is estimated on finite batches and minimized by gradient descent, which can stop at local minima, and if the model class does not contain the posterior, the guarantee says nothing about which approximation is found. Training with a strictly proper scoring rule therefore need not produce a calibrated model.

Two objectives can share the same global minimizer but have very different landscapes. In a Gaussian toy model, the energy score is much flatter than \tcvm in the direction of the posterior spread when the mean is biased, so that noisy gradients are less likely to correct the spread, whereas \tcvm penalizes miscoverage directly (Appendix~\ref{sec:toy_model}). Our experiments reflect this difference; on most of our imaging inverse problems, ES and pCRPS improve coverage but remain overconfident, whereas \tcvm reaches calibration (Section~\ref{sec:ip}).

\paragraph{Calibration objectives} Several works have proposed training objectives that encourage calibration directly, including the HPD objective of \cite{falkiewicz2023}, the rank-based statistics of \cite{yao2024simulation}, and the balanced NRE approach of \cite{delaunoy2022towards}, which favors conservative rather than exactly calibrated estimators. These methods target marginal calibration, which does not ensure calibration conditionally on the measurement \citep{pmlr-v80-kuleshov18a}. Conformal prediction \citep{angelopoulos2023conformal} instead acts after training: using a held-out calibration set, it turns the predictions of any trained model into sets that contain the true value with a chosen probability, without assumptions on the data distribution and for any calibration set size. This guarantee is also marginal, since conditional coverage is unattainable in this distribution-free setting, and the procedure does not provide a loss or score that can be used to train the model. Conditioning the calibration statistic on $y$ gives a stronger notion of calibration, explored both as a diagnostic \citep{zhao2021diagnostics} and for post-hoc recalibration \citep{dey2025towards}. Our loss enforces it during training, at the population level.

\section{The \tcvm loss: A strict divergence}
\label{sec:tarp-cvm}
 
This section builds the \tcvm loss in three stages. We first describe how the TARP coverage test of \citet{lemos2023sampling} assesses a conditional generative model: it centers a ball at a reference point chosen from the measurement, measures how much of the estimator's mass lies closer to the reference point than the true parameter does, and checks whether these masses are uniformly distributed (Section~\ref{sec:tarp-construction}). We then relate this test to credible regions and to the expected coverage probability, which places TARP alongside HPD-based tests (Section~\ref{sec:credibility}). Finally, we turn the test into a population objective and give conditions under which it vanishes only at the true posterior (Section~\ref{sec:objective}).

\subsection{The TARP construction}
\label{sec:tarp-construction}
 
TARP replaces the HPD level sets of the density (Section~\ref{sec:related}) by balls, whose mass can be estimated from samples alone. Given a pair $(\theta, y)$ drawn from the joint distribution $p(\theta, y)$, for instance a training pair $(\theta_i, y_i)$, the construction proceeds in four steps.
 
\textit{\textbf{Step 1:} choose a reference point.} For the measurement $y$, choose a reference point $\psi(y)\in\Theta$. Since this choice must be made for every measurement, we formalize it as a rule.
 
\begin{definition}[Reference rule]\label{def:rule} A reference rule is a measurable map $\psi : \mathcal{Y} \to \Theta$ that assigns to each measurement $y$ a single reference point $\psi(y)$. A rule is deterministic and specifies the reference points of all measurements jointly. Randomness over reference points enters only through a distribution $\rho$ over rules (Definition~\ref{def:loss}). \end{definition}
 
\textit{\textbf{Step 2:} construct a ball.} Center a ball at $\psi(y)$, with radius chosen so that the ball just contains the true $\theta$, that is, the ball $B\big(\psi(y), d(\theta, \psi(y))\big)$, where $d$ is a distance on $\Theta$ and $B(c, r) = \{\tilde\theta : d(\tilde\theta, c) \leq r\}$ is the closed ball.
 
\textit{\textbf{Step 3:} evaluate its mass under the estimator.} Define the \textit{credibility value} as the probability mass of this ball, where the probability is with respect to the distribution $q_\phi(\cdot \mid y)$:
$$s_\psi(\theta, y) = \mathbb{P}_{q_\phi(\tilde\theta \mid y)}\!\left[d(\tilde{\theta}, \psi(y)) \leq d(\theta, \psi(y))\right].$$
In practice, it is the fraction of draws $\tilde\theta \sim q_\phi(\cdot\mid y)$ that fall inside the ball (see Figure~\ref{fig:refs}b).

\textit{\textbf{Step 4:} evaluate coverage.} Repeating Steps 1--3 over pairs $(\theta, y) \sim p(\theta, y)$, with the same rule $\psi$, gives a distribution of credibility values. The conditional coverage of rule $\psi$ at measurement $y$ and the marginal coverage, its average over measurements, are
\begin{align}
F_\psi(\alpha \mid y) &= \mathbb{P}_{p(\theta \mid y)}\big[s_\psi(\theta, y) \le \alpha\big]\label{eq:cond-cov}
  && \text{(conditional coverage)},\\
F_\psi(\alpha) &= \mathbb{E}_{p(y)}\big[F_\psi(\alpha \mid y)\big]
  && \text{(marginal coverage)}.
\end{align}
Since $F_\psi(\alpha\mid y)$ depends on $\psi$ only through the reference point $\psi(y)$, we also write $F_c(\alpha\mid y)$ for the coverage at $y$ of balls centered at a point $c\in\Theta$, so that $F_\psi(\alpha\mid y)=F_{\psi(y)}(\alpha\mid y)$. If the estimated posterior is correct, the credibility values of the true parameters are uniformly distributed on $[0,1]$ at every measurement, $F_\psi(\alpha\mid y) = \alpha$ (Lemma~\ref{lem:post-calibrated} below). The original TARP test pools the credibility values over randomly drawn reference points. We instead compare them to the uniform distribution separately for each rule and average the resulting discrepancies (Section~\ref{sec:objective}).

\subsection{Credible regions and expected coverage}
\label{sec:credibility}

To relate TARP to HPD-based tests, we view the balls of Section~\ref{sec:tarp-construction} as credible regions of the estimator and their marginal coverage $F_\psi(\alpha)$ as an expected coverage probability.
 
\begin{definition}[Adapted from \citet{lemos2023sampling}]
    A credible region map of the estimator $\qp$ assigns to each level $\alpha\in[0,1]$ and measurement $y$, a set $R_\alpha(y)\subseteq\Theta$, such that $\Eb_{\theta\sim q_\phi(\cdot\mid y)}\big[\mathbf 1_{\theta\in R_\alpha(y)}\big] = \int_{R_\alpha(y)}q_\phi(\theta\mid y)d\theta 
    = \mathbb{P}_{\theta \sim q_\phi(\cdot | y)}[\theta \in R_\alpha(y)]
    = \alpha$.
\end{definition}
In this paper, we compare two credible region maps. The HPD region is the level set of $q_\phi(\cdot | y)$ (also the region with the smallest volume) with mass $\alpha$ \citep{box1973bayesian, hyndman1996computing}. Given a rule $\psi$, the TARP region is a ball centered at reference point $\psi(y)\in \Theta$, with radius ensuring the ball has mass $\alpha$:
$$
\begin{aligned}
    t_\alpha &= \inf\left\{ r : \mathbb{P}_{q_\phi(\theta \mid y)}
    \left(\theta \in B(\psi(y),r)\right) \geq \alpha \right\}\\
    R^\text{TARP}_{\alpha,\psi}(y) &= B(\psi(y), t_\alpha)
\end{aligned}$$
The credibility value $s_\psi(\theta, y)$
is the level at which $\theta$ enters the TARP region: $\theta \in R^\text{TARP}_{\alpha, \psi}(y) \Longleftrightarrow s_\psi(\theta, y) \leq \alpha$.
 
\begin{definition}
[Adapted from \citet{lemos2023sampling}]
\label{def:ecp}
    For any credible region map $R_{\alpha}$, the expected coverage probability at level $\alpha$ is the probability that the true parameter falls in the region:
    \begin{equation}
        \mathrm{ECP}(q_\phi, \alpha, R_{\alpha}) = \mathbb{P}_{p(\theta, y)}[\theta \in R_{\alpha}].
    \end{equation}
    We say $q_\phi$ is calibrated for $R_{\alpha}$ if $\mathrm{ECP}(\qp, \alpha, R_{\alpha})=\alpha$ for all $\alpha\in [0,1]$.
\end{definition}
 
By construction, $F_\psi(\alpha)=\mathrm{ECP}(\qp, \alpha, R^\text{TARP}_{\alpha, \psi})$. So $\qp$ is calibrated for the TARP region given by $\psi$, when $F_\psi$ is the identity, i.e., when the credibility values are uniformly distributed.
 
\begin{assumption}\label{ass:A}
$\Theta \subseteq \mathbb{R}^{d_\theta}$ is open, and for every $y$,
$p(\cdot\mid y)$ and $q_\phi(\cdot\mid y)$ have continuous densities that are strictly positive on $\Theta$.
\end{assumption}
 
\begin{lemma}\label{lem:post-calibrated}
    Under \cref{ass:A}, if $\qp(\cdot\mid y) = p(\cdot\mid y)$ for every $y$, then $F_\psi(\alpha\mid y)=\alpha$ for every rule $\psi$, for every level $\alpha$ and every $y$, and hence $F_\psi(\alpha) = \alpha$.
\end{lemma}
 
In other words, a correct estimator is always calibrated, no matter how the regions are chosen (Proof in Appendix~\ref{sec:proofs}). However the converse, calibration for a map implies $\qp=p$, depends on the choice of the map, and is the central question of Section~\ref{sec:objective}.

\subsection{The population objective}\label{sec:objective}
 
\begin{definition}\label{def:loss} For a distribution $\rho$ over reference rules, we introduce the loss function
    \begin{equation}\label{eq:loss}
        \Dr(p,\qp) = \Eb_{\psi\sim\rho}\int_0^1  \left(F_\psi(\alpha) - \alpha\right)^2 d\alpha.  
    \end{equation}
\end{definition}
$\Dr=0$ iff the estimator is calibrated for every rule $\psi$ in the support of $\rho$. 
For a single rule, the integrand compares each credibility level $\alpha$ with the frequency $F_\psi(\alpha)$ at which the true parameter falls in the corresponding region. The inner integral is thus a continuous analogue of the expected calibration error used in classification \citep{naeini2015obtaining}, and closely resembles the calibration error of \citet{pmlr-v80-kuleshov18a} for regression. The outer expectation over rules is what distinguishes it from these per-model calibration scores, and the choice of $\rho$ determines the properties of $\Dr$. We motivate the requirements on $\rho$ below and formalize them in \cref{def:adm}.
 
For a fixed rule $\psi$, the integral is the Cramér--von Mises distance \citep{Cramer01011928} $D_{\rm CvM}(X_1,X_2)=\int_0^1\big(\mathbb P[X_1\le\alpha]-\mathbb P[X_2\le\alpha]\big)^2d\alpha$, where  $X_1=s_\psi(\theta,y)$, the credibility of the true parameters, whose CDF is $F_\psi$, and $X_2\sim\mathcal{U}[0,1]$ is the uniform distribution. $\mathcal D_\rho$ is non-negative and, under \cref{ass:A} and for any admissible
$\rho$ (Definition~\ref{def:adm}), vanishes if and only if $q_\phi=p$ (Theorem~\ref{thm:proper}); in this sense
it is a divergence between conditional distributions. It is not a metric, since it
is not symmetric in $(p,q_\phi)$ and does not satisfy the triangle inequality, and it
is not a scoring rule, since it is quadratic in the distribution of $(\theta,y)$
rather than an average of per-example scores.

\paragraph{Constant reference points cannot identify the posterior.}
A rule that ignores the measurement lets the prior pass the coverage test. Let us consider a constant rule $\psi(y)=c$. If our estimator returns the prior, $\qp(\theta\mid y)=p(\theta)$, the TARP region $R^\text{TARP}_{\alpha,\psi}(y)=R^\text{TARP}_{\alpha,c}$ no longer depends on $y$, since both its center and the distribution defining its radius are fixed. Then
$\mathrm{ECP}(q_\phi, \alpha, R^\text{TARP}_{\alpha, \psi}) = \Eb_{p(\theta,y)}\big[\mathbf 1_{\theta\in R^\text{TARP}_{\alpha, c}}\big] = \int p(y)\int_{R^\text{TARP}_{\alpha, c}} p(\theta\mid y)\,d\theta\, dy=\int_{R^\text{TARP}_{\alpha, c}} \left(\int p(\theta\mid y)p(y)\, dy\right)d\theta = \int_{R^\text{TARP}_{\alpha, c}} p(\theta)\,d\theta$.. Since $\qp(\theta\mid y)=p(\theta)$ and by definition of a credible region, we get $\mathrm{ECP}(q_\phi, \alpha, R^\text{TARP}_{\alpha, \psi})=\alpha$, which evaluates the estimator as calibrated. 

\paragraph{A single measurement-dependent rule is still insufficient.}
Reference points must depend on the measurement, but this alone does not guarantee that the loss identifies the posterior. A single rule gives each measurement its own reference point, but its coverage $F_\psi(\alpha)=\int F_\psi(\alpha\mid y)\,p(y)\,dy$ is an average over measurements, and errors at different measurements can cancel. 
Suppose there are two measurements $y_0$ and $y_1$, each with probability $\frac12$, and an estimator that is too wide at $y_0$ and too narrow at $y_1$, with $F_\psi(\alpha\mid y_{\iota})=\alpha+(-1)^\iota \delta\,\alpha(1-\alpha)$ for $\iota \in \{0,1\}$ and $\delta\in(0,1)$. The coverage curves at $y_0$ and $y_1$ deviate from the diagonal in opposite directions and average to it, so $F_\psi(\alpha)=\alpha$ at every level: the estimator is calibrated under $\psi$ yet wrong at both measurements. Choosing a better single rule does not help, since the counterexample holds for any $\psi$.

\paragraph{Comparing rules requires local differences.}
For a wrong estimator, the conditional coverage $F_\psi(\alpha\mid y)$ (Equation~\ref{eq:cond-cov}) depends on where the ball is centered, so errors can be revealed by comparing different reference points at the same measurement. The loss, however, only sees marginal coverages $F_\psi(\alpha)$, so this comparison must be made across several rules. It isolates a subset of measurements only if the collection contains pairs of rules that agree on most measurements and differ on that subset: the difference between their marginal coverages then depends only on the subset. If every pair of rules differs at all measurements, errors can still cancel within each marginal average.
 
For instance, a simpler reference would add noise to a point estimate $b(y)$ at scale $\varsigma$ (see Section~\ref{sec:reference-points} for definitions), $\psi(y_i) = b(y_i) + \varsigma\xi_i$, with $\xi_i$ drawn either independently for each example or once for the whole batch. Neither is sufficient. 
With independent draws, the batch yields a single coverage curve, $\bar F(\alpha) = \int \mathbb{E}_{\xi}\big[F_{b(y)+\varsigma\xi}(\alpha \mid y)\big]\, p(y)\, dy$, and a single statistic $\int_0^1 \big(\bar F(\alpha) - \alpha\big)^2 \, d\alpha$. The expectation over reference points then sits inside the square of Equation~\ref{eq:loss}, and the examples where $F_{b(y_i)+\varsigma\xi_i}(\alpha \mid y_i) > \alpha$ can balance those where $F_{b(y_i)+\varsigma\xi_i}(\alpha \mid y_i) < \alpha$, so that $\bar F(\alpha) = \alpha$ even if $F_{b(y_i)+\varsigma\xi_i}(\alpha \mid y_i) \neq \alpha$ for every example.
If instead a single displacement $\xi_r$ is shared by the batch, each draw is a rule $\psi_r(y) = b(y) + \varsigma\xi_r$ giving a statistic $I_{\psi_r}$, and the rules satisfy (R1) below. But each $I_{\psi_r}$ constrains $F_{\psi_r}(\alpha \mid y)$ only through an average over all measurements, so nothing prevents over-coverage at some measurements from balancing under-coverage at others for every $\xi_r$ at once, and Theorem~\ref{thm:proper} does not apply to these rules.

\paragraph{Patching rules resolves the problem.}
Given two reference rules, a patched rule uses the first rule for measurements on one side of a hyperplane and the second rule for measurements on the other side. Thus, we can change the reference points for one subset of measurements while leaving the rest unchanged. In the example above, with $\delta=0.8$ and $\alpha=\frac12$, the coverages are $0.7$ and $0.3$ under $\psi$;  take a second calibrated rule $\psi'$, centered elsewhere, with coverages $0.6$ and $0.4$. Switching rules across measurements, $\psi$ at $y_0$ and $\psi'$ at $y_1$, gives $\frac12(0.7+0.4)=0.55$, and the miscalibration appears.
 
\begin{definition}\label{def:patch}
For rules $\psi,\psi'$ and a measurable set $H\subseteq\mathcal Y$, the patched rule uses $\psi$ on $H$ and $\psi'$ elsewhere, $(\psi\oplus_H\psi')(y)=\psi(y)\,\mathbf 1_{y\in H}+\psi'(y)\,\mathbf 1_{y\notin H}$.
\end{definition}
 
Patching along all half-spaces is enough (Lemma~\ref{lem:loc}), which is condition (R2) below. The reference rules must have two properties. First, at every measurement, the available reference points must be sufficiently rich to probe the entire parameter space. Second, the collection of rules must allow reference points to be changed on arbitrary half-spaces of measurement space while remaining unchanged elsewhere. These two properties ensure that coverage errors cannot be hidden by averaging, and that the resulting coverage tests distinguish different conditional distributions.

\begin{definition}\label{def:adm}
A distribution $\rho$ over rules is \emph{admissible} if there exists a countable family of rules $\Psi\subseteq\operatorname{supp}\rho$ (Appendix~\ref{sec:support}) such that
 
\begin{itemize}
    \item[(R1)] For every $y$, the centers
        $\{\psi(y):\psi\in\Psi\}$ are dense in $\Theta$.
  \item[(R2)] 
  For all $\psi,\psi'\in\Psi$ and every
        half-space $H=\{y:\langle u,y\rangle\le t\}$, with $u$ a unit vector of $\mathbb R^{d_y}$ and $t\in\mathbb R$, the patched rule $\psi\oplus_H\psi'$ is
        in $\operatorname{supp}\rho$.
\end{itemize}
\end{definition}
 
With (R1), the loss can place reference points anywhere near the posterior mass at each measurement. With (R2), they may be used on the two sides of a hyperplane in measurement space, which is enough to tell measurements apart.
 
The switch rule of Equation~\ref{eq:reference-points} avoids both problems of the noisy point estimate above. 
Because each side of the hyperplane shares one displacement, each draw is a rule with its
own statistic. Using one displacement on $H$ and another on its complement changes the
reference points of the measurements in $H$ only, so errors inside $H$ can no longer be
balanced by errors outside it. Calibration of every such rule, for every $H$, forces the
coverage at almost every measurement to be the same for every displacement
(condition (R2) and Lemma~\ref{lem:loc}).

\begin{definition} The \tcvm loss is \autoref{eq:loss} for an admissible $\rho$.    
\end{definition}
 
\begin{theorem}[TARP--CvM is a strict divergence]\label{thm:proper}
Under \cref{ass:A} and for any admissible $\rho$, $\Dr(p,\qp)=0$ if and only if $\qp(\cdot\mid y)=p(\cdot\mid y)$ for almost all $y$.
\end{theorem}
 
Proof in Appendix~\ref{sec:proofs}. 
The direction $\qp = p \Rightarrow \Dr = 0$ follows from Lemma~\ref{lem:post-calibrated}.
The converse proceeds in three steps. First, zero loss implies calibration for every rule in the support of $\rho$. Second, patching (R2) implies that
the conditional coverage at almost every measurement does not depend on the reference
point. Third, since reference points are dense (R1), this invariance holds only if the
estimated and true posteriors agree.
 
\citet[][Theorem 3]{lemos2023sampling} show that a posterior estimator is exact if its expected coverage is correct for every function assigning reference points to measurements. Their argument relies on a functional derivative with respect to this function. They do not specify how reference points should be selected from the measurement, and in their experiments reference points are drawn independently of it, a setting in which the prior also passes the test. Moreover, the original TARP pools coverage over randomly drawn reference points, and the pooled coverage can match the nominal level even when a distribution is miscalibrated with respect to an individual rule, because their errors can cancel in the average. Theorem~\ref{thm:proper} shows how our \tcvm addresses both issues. It gives an explicit sufficient condition on how reference points depend on the measurement (Definition~\ref{def:adm}). Our loss averages the miscalibration of a distribution with respect to each rule rather than pooling, so that zero loss implies calibration for every rule in the support.
 
\section{Making the objective trainable}

At each training iteration, we draw a batch of paired observations and ground-truth signals $(\theta_i,y_i)$, generate samples for each observation \(\tilde \theta_{i,m}\), and construct several reference rules. For each rule, we compute the credibility of every ground-truth signal \(\theta_i\) relative to the corresponding generated samples. We then calculate a Cramér–von Mises discrepancy between the empirical distribution of these credibility values and the uniform distribution. The average discrepancy over reference rules is added to the base generative-model training loss.

\subsection{The closed form estimator}

The objective term in \autoref{eq:loss} is the squared area $I_\psi = \int_0^1\left(F_\psi(\alpha)-\alpha \right)^2d\alpha$ between the cumulative distribution function of the credibility values and the diagonal. $F_\psi$ is empirically estimated with $N$ credibility values $s_i=s_\psi(\theta_i,y_i)$ from pairs $(\theta_i,y_i)$
of the batch, $\hat F_\psi(\alpha)\approx\frac{1}{N}\sum_i \mathbf 1_{s_i\leq\alpha}$, computed with the same rule $\psi$.
$I_\psi$ is a Cram\'er-von Mises statistic, and its derivative admits a closed-form expression \citep{yao2024simulation}.
The statistic measures the squared discrepancy between the empirical distribution of the credibility values and the uniform distribution. The usual integral representation involves indicator functions and is not convenient for gradient-based optimization. By integrating analytically over the credibility levels, we obtain the following expression in terms of the sorted credibility values.
See Appendix~\ref{sec:app-unbiased-estimator}
for details on deriving this empirical estimator:
\begin{equation}\label{eq:cvm}
\cvm(s_{1:N}) = \frac{1}{N}\sum_i s_i^2
- \frac{1}{N^2}\sum_{k=1}^N (2k-1)\,s_{(k)}
+ \frac{1}{3},
\end{equation}
where $s_{(1)}\le\dots\le s_{(N)}$ are the sorted credibility values. Writing $\operatorname{rank}(i)$ for the position of $s_i$ in this ordering, so that $s_i = s_{(\operatorname{rank}(i))}$, the gradient is
\begin{equation}
\frac{\partial\cvm}{\partial s_i}=\frac{2s_i}N-\frac{2\,\mathrm{rank}(i)-1}{N^2}.
\end{equation}
The closed form expression is what makes the statistic differentiable, by turning evaluation on a grid of levels $\alpha$, where most indicator counts would have zero gradient, into a polynomial of the sorted values. Integrating over $\alpha$ first removes the indicators, and the resulting expression depends on $s_i$ values and rank. It was used by \citet{yao2024simulation} to turn a rank-uniformity test into a differentiable objective for simulation-based inference. We differ in the choice of the statistic. Coordinate-wise ranks constrain only the one-dimensional marginals of the estimator, so the prior estimator also passes the test, whereas uniformity of the credibility value across admissible rules forces the estimator to equal the posterior as stated in Theorem~\ref{thm:proper}.

\paragraph{Unbiased estimator} \autoref{eq:cvm} is the plug-in estimate of the inner integral of \autoref{eq:loss}.  Written as an average over all pairs $(i,j)$ of credibility values, it includes the diagonal pairs $i=j$, so its expectation exceeds it by $\frac1N\int_0^1F_\psi(1-F_\psi)\,d\alpha$, a term that does not vanish at the posterior. Excluding the diagonal pairs gives an unbiased U-statistic \citep{Hoeffding},
\begin{equation}\label{eq:cvm-u}
\cvm_U(s_{1:N})=\frac1N\sum_i s_i^2-\frac{2}{N(N-1)}\sum_{k=1}^N(k-1)\,s_{(k)}+\frac13.
\end{equation}
\autoref{eq:cvm-u} is therefore preferable when an unbiased estimate of the population discrepancy is desired (\cref{prop:cvm-u}, see Appendix~\ref{sec:app-unbiased-estimator} for derivation). In our experiments, however, we use the simpler plug-in statistic in Equation~\ref{eq:cvm}. Its finite-sample bias is not removed, but the same estimator is used consistently throughout training.

\subsection{Reference points}\label{sec:reference-points}

\begin{figure}[t]
    \centering
    \includegraphics[width=0.7\linewidth]{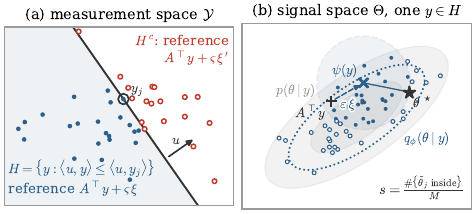}
    \caption{\textbf{The reference sampler.} (a) 
    In measurement space, a hyperplane with random direction $u$ through a random measurement $y_j$ splits the batch: measurements on the shaded side (\mkfilled) receive the displacement $\xi$, those on the other side (\mkred) receive $\xi'$. (b) In signal space, for one measurement on the $\xi$ side. The adjoint reconstruction $A^\top y$ ($\boldsymbol{+}$), the reference $\psi(y)=A^\top y+\varsigma\xi$ ($\color{figblue}{\times}$), the truth $\theta^*$ ($\star$), and the ball centered at $\psi(y)$ with radius $d(\psi(y), \theta^*)$. Estimator draws inside the ball are filled (\mkfilled), draws outside are hollow (\mkhollow); the credibility value $s$ is the filled fraction. Grey ellipses are the posterior, the dotted ellipse the overconfident estimator.}\label{fig:refs}
\end{figure}

The objective \autoref{eq:loss} is an average over reference rules. Each optimization step draws $R$ rules and applies each one to every pair in the batch. It then computes the CvM statistic of \autoref{eq:cvm} separately for each rule and averages the $R$ statistics. This keeps the expectation over rules outside the squared deviation, as in \autoref{eq:loss}. The rules are drawn from a mixture of three groups: switch rules, prior rules and posterior rules.

\paragraph{Switch rules.}
\autoref{fig:refs} illustrates the construction. A switch rule needs a point estimate $b(y)$ with nontrivial posterior likelihood. For a linear forward model $y=A\theta+\varepsilon$, we take the adjoint reconstruction $b(y)=A^\top y$, or the pseudo-inverse $A^+y$. Without an adjoint, $b$ can be a regressor of the posterior mean, trained beforehand on the same pairs (\cref{sec:sbi}). A switch rule splits the measurements of the batch with a hyperplane and displaces the point estimate by one displacement on one side and by another on the other side, so that it is a single rule patched across a half-space (Definition~\ref{def:patch}). It is determined by two displacements $\xi,\xi'$, a unit vector $u$ in measurement space that orients the hyperplane, and an example $j$ of the batch through which it passes, and assigns to every example $y_i$ the reference point
\begin{equation}\label{eq:reference-points}
    \psi(y_i) = b(y_i) + \varsigma \left( \xi\, \mathbf{1}_{\langle u, y_i \rangle \leq \langle u, y_j \rangle} + \xi'\, \mathbf{1}_{\langle u, y_i \rangle > \langle u, y_j \rangle} \right),
\end{equation}
where $\varsigma\in\mathbb{R}^{d_\theta}_{>0}$ is a fixed reference scale, applied component-wise and treated as a constant. Drawing $\xi,\xi'\sim\mathcal N(0,I)$, a random unit vector $u$ and a random example $j$ chooses a switch rule at random; once drawn, it is a fixed rule in the sense of Definition~\ref{def:rule}.
In our experiments, we set $\varsigma$ to the root-mean-square residual of $b$ on a held-out tenth of the training simulations,

\begin{equation}\label{eq:scale}
\varsigma = \big(\mathbb{E}\big[(b(y) - \theta)^{2}\big]\big)^{1/2},
\end{equation}
with the square and root taken element-wise, so that the reference displacements match the average posterior scale. $b$ and $\varsigma$ are both frozen before the posterior estimator is trained. 

The hyperplane splits the batch into two groups that receive different displacements, so within a rule each measurement receives one reference point. Across the R rules of a step, each measurement receives R reference points. The switch rules satisfy both admissibility conditions in \cref{def:adm}: their reference points are dense in parameter space, and they can implement arbitrary half-space patches. 
The displacements have positive densities, so reference points come arbitrarily close to every point of $\Theta$ (R1). Every rule is a patch of two rules across a half-space, and every patch can be drawn, since the displacements, the direction and the position of the hyperplane are all drawn from distributions with full support (R2).

\paragraph{Prior and posterior rules.} Admissibility guarantees that the true posterior is the unique minimizer of the loss, but we observe that adding other rules helps training. A prior rule uses a single reference point drawn from the prior and shared by the batch. A posterior rule uses the $M$ posterior draws of each pair as reference points, detached from the computational graph. For the reference $\tilde \theta_m$, the credibility of $\theta_i$ is computed from the $M-1$ other draws. When a generated posterior sample is used as the reference point, that same sample must be excluded from the set of samples used to estimate the credibility value. Otherwise, the reference point would contribute a guaranteed zero-distance sample, changing the distribution of credibility values even when the estimator is correct.

\paragraph{\cref{thm:proper} still applies.} Every term of the loss is nonnegative, and the switch rules keep a fixed share of the mixture, so the loss can only vanish if the loss restricted to switch rules vanishes, which forces $\qp=p$. The switch rules are essential for the theoretical guarantee, and the other two are supplementary choices intended to help optimization.

\subsection{Relaxation of the indicator}

To differentiate the loss with respect to $\phi$, we backpropagate through the sampler. For a diffusion model, we draw all the noises of the sampler in advance and hold them fixed, so that the sampler becomes a deterministic function of $\phi$. This is the reparameterization trick \citep{kingma2013auto}. The exact credibility value is the fraction of generated samples whose distance from the reference point is no greater than the distance of the true signal. The indicator in this definition is nondifferentiable, so we replace it with a sigmoid approximation $\sigma$ at temperature $\tau>0$, for $\tilde\theta_m\sim q_\phi(\cdot\mid y)$,
\begin{equation}\label{eq:relax}
    s_\psi(\theta,y)\approx\frac{1}{M}\sum_{m=1}^M
    \sigma\!\left(\frac{d\big(\theta,\psi(y)\big)-d\big(\tilde\theta_m,\psi(y)\big)}{\tau}\right). 
\end{equation}
Algorithm~\ref{alg:tcvm} summarizes one training step, and Appendix~\ref{sec:appendix-algo} details its complexity.

\begin{algorithm}[t]
\caption{One training step with the \tcvm regularizer}
\label{alg:tcvm}
\begin{algorithmic}[1]
\REQUIRE batch $\{(\theta_i, y_i)\}_{i=1}^N$, posterior point estimator $b:\mathcal{Y}\to\Theta$, reference scale $\varsigma$ (Equation~\eqref{eq:scale}), base loss $\mathcal{L}_\text{base}$ of the generative model, weight $\lambda$, number of draws $M$, numbers of switch, prior and posterior rules $R_s, R_p, R_\text{post}$ with $R = R_s + R_p + R_\text{post}$, temperature $\tau$
\FOR{$i = 1, \dots, N$}
  \STATE Draw $\tilde\theta_{i,1}, \dots, \tilde\theta_{i,M} \sim q_\phi(\cdot \mid y_i)$ with the reparameterization trick
\ENDFOR
\FOR{$r = 1, \dots, R$}
  \STATE Let $k \in \{\text{switch}, \text{prior}, \text{posterior}\}$ be the group of rule $r$
  \STATE Draw a rule $\psi_r \sim \rho_k$:
  \IF{$k = \text{switch}$}
    \STATE Draw $\xi, \xi' \sim \mathcal{N}(0, I)$, a unit vector $u$ and an index $j$
    \STATE $\psi_r(y) = b(y) + \varsigma\big(\xi\,\mathbf{1}_{\langle u, y\rangle \le \langle u, y_j\rangle}
      + \xi'\,\mathbf{1}_{\langle u, y\rangle > \langle u, y_j\rangle}\big)$
    \STATE $\mathcal{M}_r \gets \{1, \dots, M\}$
  \ELSIF{$k = \text{prior}$}
    \STATE Draw $c \sim p(\theta)$ and set $\psi_r(y) = c$
    \STATE $\mathcal{M}_r \gets \{1, \dots, M\}$
  \ELSE
    \STATE Draw an index $m'$ and set $\psi_r(y_i) = \mathrm{sg}(\tilde\theta_{i,m'})$
    \STATE $\mathcal{M}_r \gets \{1, \dots, M\} \setminus \{m'\}$
  \ENDIF
  \FOR{$i = 1, \dots, N$}
    \STATE $s_i \gets \frac{1}{|\mathcal{M}_r|} \sum_{m \in \mathcal{M}_r}
      \sigma\big(\big[d(\theta_i, \psi_r(y_i)) - d(\tilde\theta_{i,m}, \psi_r(y_i))\big]/\tau\big)$
  \ENDFOR
  \STATE $\mathcal{L}_r \gets \mathrm{CvM}(s_{1:N})$ \hfill $\triangleright$ Equation~\eqref{eq:cvm}
\ENDFOR
\STATE Compute $\mathcal{L}_{\text{base}}(\phi)$ on its own batch \hfill $\triangleright$ e.g.\ DSM or NLL
\STATE $\mathcal{L} \gets \mathcal{L}_{\text{base}}(\phi) + \lambda \frac{1}{R}\sum_{r=1}^R \mathcal{L}_r$
\STATE Update $\phi$ with a gradient step on $\mathcal{L}$
\end{algorithmic}
\end{algorithm}

\section{Experiments}

\subsection{\tcvm as training signal}
\label{sec:fig1}

We first consider a setting in which the true posterior is known analytically. This experiment isolates the distinction between marginal coverage and posterior accuracy, allowing us to directly compare the minimizers of the HPD and \tcvm objectives. We illustrate \autoref{thm:proper} on a Gaussian linear inverse problem $y=A\theta+\varepsilon$, with a Gaussian prior and fewer measurements than parameters (Appendix~\ref{sec:appendix-toy}). The posterior is Gaussian, $p(\theta\mid y)=\mathcal N(W^*y,S^*)$, and we fit a Gaussian estimator $\qp(\theta\mid y)=\mathcal N(Wy,S)$ by gradient descent on either an HPD-loss or \tcvm. The estimator is HPD-calibrated for every $W$, provided that $S=S^*+(W^*-W)\Sigma_y(W^*-W)^\top$, where $\Sigma_y$ is the covariance of the measurements. Thus, an estimator with a biased conditional mean can still achieve correct HPD coverage if its covariance is sufficiently inflated. The HPD objective therefore admits a family of minimizers, including the prior, even though most of these distributions do not equal the true posterior. This manifold of minimizers contains the prior, at $W=0$, and every point of it other than the posterior has a biased mean and a covariance wider than the posterior's. Gradient descent on HPD coverage can therefore stop at any point of this manifold, not only at the posterior. With measurement-dependent references, \tcvm instead has the posterior as its unique minimizer (\autoref{thm:proper}). In \autoref{fig:fig1}, \tcvm recovers the posterior contours, whereas the HPD fit is biased and too conservative. The HPD coverage error cannot distinguish the two models, although their KL divergence to the posterior differs strongly.

\subsection{Imaging inverse problems}\label{sec:ip}

\begin{figure}[h]
    \centering
    \includegraphics[width=0.5\linewidth]{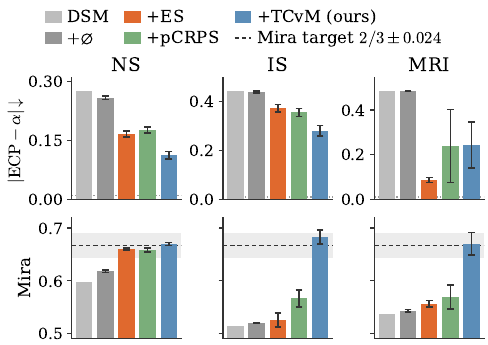}
    
    \caption{Coverage before and after calibration.}
    \label{fig:tab:coverage}
\end{figure}

We study posterior coverage calibration in high-dimensional posteriors on three inverse problems from \textsc{InverseBench} \citep{zheng2025inversebench}. Each has the form $y = \mathcal{A}(\theta) + \varepsilon$. These are \textit{linear inverse scattering} (IS) and \textit{compressed-sensing MRI} (MRI), whose forward operators are linear, and \textit{Navier–Stokes} (NS), whose forward operator is nonlinear and only available as a black-box solver (Appendix~\ref{sec:appendix-ip}). For IS, $\varepsilon$ is additive Gaussian noise. For MRI and NS, $\varepsilon\equiv0$, so the posterior has no density on the full signal space. On MRI, \tcvm therefore measures distances in the null space of $A$, where \cref{thm:proper} applies (Appendix~\ref{sec:appendix-ip}).

Unlike the plug-and-play methods \textsc{InverseBench} was designed for, we train fully amortized conditional models on pairs $(\theta, y)$. The measurement is concatenated to the network input, as its adjoint reconstruction $\mathcal{A}^\top y$ for IS and MRI, and at its native $64\times64$ resolution for NS, which matches the resolution of the signal after $2\times2$ patching. All models are pretrained with denoising score matching \citep[DSM,][]{song2021scorebased}. They are then fine-tuned for 24k steps with one of four objectives, DSM alone ($\varnothing)$, a control that isolates the effect of additional training, or DSM plus a calibration regularizer, which is the energy score (ES), the pixel-wise CRPS (pCRPS), or \tcvm. All three regularizers use the same calibration batch, the same number of posterior samples per measurement, and the same $T=20$-step DDIM sampler \citep{song2021denoising}, through which gradients are backpropagated (see Appendix~\ref{sec:appendix-ip} for details).

We measure coverage with the $L^1$ TARP coverage error, $\int_0^1 |\mathrm{ECP}(q_\phi, \alpha, R^\text{TARP}_{\alpha, \psi}) - \alpha|\,d\alpha$. We compute the coverage using standard binary indicators rather than the sigmoid relaxations used during training, and use the $\ell_2$ distance to define the TARP balls.
We also use the Mira score \citep{sharief2026mira}, which equals $2/3$ for a calibrated model; values below the null band indicate overconfidence. For quality, we report in Table~\ref{tab:quality} the PSNR, SSIM, and relative $L^2$ error of the posterior mean over $K = 32$ samples, the PSNR of a single draw, and the posterior spread (the per-pixel standard deviation across the $K$ samples, in physical units), all computed on 64 held-out observations, together with the DSM loss on the full validation split, averaged over three seeds.

All pretrained models are strongly overconfident (Mira between 0.51 and 0.61, Figure~\ref{fig:tab:coverage}). Additional DSM training does not fix this, so the gains below come from the regularizers. On NS and IS, \tcvm gives the lowest coverage error (0.11 and 0.28, against 0.17 and 0.35 for the best scoring rule). On NS all three regularizers reach the Mira null band, but on IS only \tcvm does (0.68, against 0.53 for ES and 0.57 for pCRPS). On MRI, the two metrics disagree. ES has the lowest coverage error, but remains overconfident according to Mira, while \tcvm is the only method inside the band.

Increasing posterior variance can reduce the PSNR of individual posterior draws even when the posterior mean remains unchanged. This is not necessarily a deterioration in reconstruction accuracy: it can reflect the fact that the model is representing uncertainty that was previously missing. Calibration has little effect on the quality of generated samples. The validation DSM loss stays within $1\%$ of the control for every method. Posterior-mean PSNR is unchanged on IS and NS and drops by $0.8$ dB on MRI. Per-draw PSNR drops more, by $0.35$ to $1.9$ dB, but this reflects the metric rather than accuracy. The expected squared error of a draw $\tilde\theta \sim q(\cdot\mid y)$ decomposes as $\mathbb{E}\big[\|\tilde\theta-\theta^*\|^2\big] = \big\|\mathbb{E}_q[\tilde\theta]-\theta^*\big\|^2 + \mathrm{Var}_q(\tilde\theta)$, where $\mathrm{Var}_q(\tilde\theta) = \operatorname{tr}\mathrm{Cov}_q(\tilde\theta)$ denotes the total variance, so a model that collapses its variance gains per-draw PSNR without improving its mean. Correct coverage requires nonzero variance, and \tcvm widens the posterior by $15$ to $58\%$ relative to the control, which lowers per-draw PSNR accordingly. On NS, the scoring rules instead improve the posterior mean by about 1 dB while slightly reducing the spread. This suggests that the pretrained model had errors due to both a biased mean, which the scoring rules correct, and insufficient spread, which \tcvm corrects. The wider spread of \tcvm may also partly compensate for residual error in the posterior mean.

\begin{figure}[h]
    \centering
\includegraphics[width=0.6\linewidth]{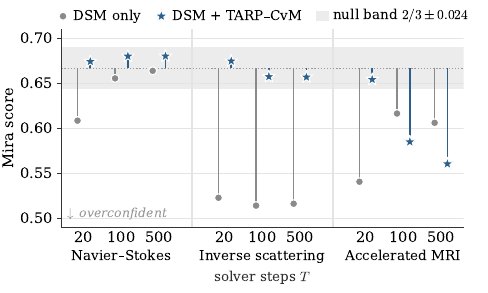}
\caption{Coverage under varying sampler depth. \tcvm was trained at $T=20$.}
    \label{fig:mira-t-sweep}
\end{figure}

\paragraph{Sampler depth.} Because the calibration term backpropagates through the sampler, training fixes the sampler depth, but any depth can be used at inference. We train with $T = 20$ and evaluate at $T \in \{20, 100, 500\}$ (Figure~\ref{fig:mira-t-sweep}). On NS and IS, Mira stays inside the null band at all depths, so calibration learned with a short chain survives at longer ones. 
On MRI, however, Mira decreases and leaves the null band at higher $T$. At T = 100 and 500 it even falls below the DSM-only model, whose Mira improves with T. This suggests that on MRI part of the learned calibration compensates for the discretization error of the 20-step sampler rather than correcting the underlying score, so the correction no longer matches a finer discretization. In practice, the model should be sampled at the depth it was calibrated for, or calibrated at the depth used for inference, at a training cost that grows linearly with $T$.

\subsection{SBI benchmark}\label{sec:sbi}

In contrast to the imaging experiments, the SBI benchmarks are low-dimensional problems for which standard NPE is already relatively well calibrated. Our objective is to establish that \tcvm can match a density-based calibration method while requiring only posterior samples. We evaluate our method on four problems from the SBI benchmark introduced in \citet{hermans2022a}, namely \textit{SLCP, Weinberg, Spatial SIR and Lotka–Volterra}, each at three simulation budgets $\{2^{10}, 2^{13}, 2^{16}\}$. We compare conditional normalizing flows trained with Neural Posterior Estimation \citep[NPE,][]{papamakarios2016fast, lueckmann2017flexible,deistler2022truncated, pmlr-v97-greenberg19a}, with the HPD-based regularizer of CalNPE \citep{falkiewicz2023}, and with \tcvm. In SBI, we usually only have access to samples from the simulator, and cannot evaluate the adjoint operator of the forward model. In order to get a point estimate $b(y)$ close to the posterior for reference points when only pairs of $(\theta, y)$ are available, we train a multilayer perceptron (MLP) network targeting the posterior mean $y\mapsto\mathbb{E}_{p(\theta\mid y)}[\theta]$, with a mean-square-error loss, and freeze it during calibration of the posterior generative model. All methods share the same training protocol, with the same simulation budgets, validation set and checkpoint selection (Appendix~\ref{sec:appendix-sbi}).

Under this matched protocol, NPE is already nearly calibrated on these low-dimensional problems. 
On the two TARP coverage maps, log posterior density and joint MMD on $(\theta, b(y))$, a single method is significantly better than the other two in only two of the 48 comparisons: \tcvm on the switch map for Weinberg at $2^{10}$, and CalNPE on log density for Spatial SIR at $2^{10}$ (Table~\ref{tab:sbi}, Appendix~\ref{sec:appendix-sbi}). 

\textit{\tcvm thus matches CalNPE using only posterior samples, without evaluating the density.}
Its log density does not differ significantly from that of NPE in any setting, but is lower than that of CalNPE on Spatial SIR at the two smaller budgets. CalNPE achieves a lower HPD coverage error, the statistic it optimizes, with a significant difference in four settings. On the independent Mira score, \tcvm is closest to the calibrated value 2/3 in half of the settings, and CalNPE deviates most from it in 10 of 12 settings, consistent with its conservativeness objective, a one-sided penalty on under-coverage that leaves over-coverage unpenalized, which we used to match the configuration reported by \citet{falkiewicz2023}. None of these differences is significant.

\section{Conclusion}
We introduced \tcvm, a differentiable coverage loss that vanishes if and only if the learned conditional distribution equals the true posterior. It requires only samples from the model, applies to any generative architecture, and is a strict divergence (Theorem~\ref{thm:proper}). On SBI benchmarks, it matches a density-based coverage loss with the same number of model evaluations, without evaluating the density. On imaging inverse problems, including one with a nonlinear black-box forward model, it is the only regularizer that brings the posterior into the Mira null band on all three problems, with little change in the quality of point estimates. The main practical design choice is the construction of reference points, which are built around a point estimate $b(y)$. For linear forward models, the adjoint $A^\top y$ works well. Without an adjoint, an auxiliary regressor of the posterior mean can be trained at extra cost, as in our SBI experiments. Future work includes other nonlinear inverse problems, such as full waveform inversion and black hole imaging, and SBI problems in cosmology, where the forward model is a physics simulator and posteriors are often multimodal.

\newpage
\section*{AI use statement}

In this work, we used generative AI tools for formulating mathematical claims, assisting in the writing and checking of proofs, providing feedback on the design of the method and experiments, and assisting in implementing the method and experiments. Additionally, we used generative AI tools to edit and rewrite the manuscript for clarity.
We have reviewed all AI-assisted work. All theorems and proofs were checked line by line by the authors. AI-generated code was reviewed and tested by the authors. Results were interpreted and conclusions drawn by the authors, and all edited text was reviewed by the authors. We take responsibility for the final content of this work, including text, claims, or artifacts produced with the aid of generative AI.

\section*{Acknowledgments}
All authors gratefully acknowledge the support of the NSF-Simons AI-Institute for the Sky (SkAI) via grants NSF AST-2421845 and Simons Foundation MPS-AI-00010513. 
RW further gratefully acknowledges the support of the NSF-Simons National Institute for Theory and Mathematics in Biology (NITMB), via grants NSF DMS-2235451 and Simons Foundation MP-TMPS-00005320, and AFOSR FA9550-18-1-0166.
This research used the DeltaAI advanced computing and data resource, which is supported by the National Science Foundation (award OAC 2320345) and the State of Illinois. DeltaAI is a joint effort of the University of Illinois Urbana-Champaign and its National Center for Supercomputing Applications.

\bibliography{references}

\clearpage
\appendix
\thispagestyle{empty}

\onecolumn

\section{Proofs}\label{sec:proofs}

This section proves Theorem~\ref{thm:proper}. Section~\ref{sec:support} defines the support of a distribution over rules, Section~\ref{sec:conventions} recalls the notation, and Section~\ref{sec:prelim} collects preliminary facts: the regularity of the radial mass function, the calibration of the true posterior for every rule, and a Cramér--Wold argument showing that integrals over half-spaces determine a function of the measurement. The direction ``$q_\phi=p$ implies $\mathcal D_\rho=0$'' holds for any $\rho$ and follows from Lemma~\ref{lem:post-calibrated}. The converse is proved in three steps, one per subsection:
\begin{enumerate}[label=(\roman*)]
    \item if $\mathcal D_\rho=0$, every rule in the support of $\rho$ is calibrated (Section~\ref{sec:step-support})
    \item by patching (R2), the coverage at each measurement then does not depend on the reference point (Section~\ref{sec:step-invariance})
    \item since reference points are dense (R1), this invariance forces $p(\cdot\mid y)=q_\phi(\cdot\mid y)$ (Section~\ref{sec:step-densities}).
\end{enumerate}
Section~\ref{sec:step-theorem} assembles the three steps.

\subsection{Rules, their topology and the support of \texorpdfstring{$\rho$}{rho}}\label{sec:support}
 
A reference rule is a measurable map $\psi:\mathcal Y\to\Theta$. We define how far apart two rules $\psi$ and $\psi'$ are by
\[
  d_0(\psi,\psi')=\Eb_{y\sim p(y)}\big[\min\{1,\|\psi(y)-\psi'(y)\|\}\big],
\]
the metric of convergence in probability. Two rules are close if their reference points are close for most measurements. The \textit{support} of a distribution $\rho$ over rules is the set of rules that
$\rho$ approaches arbitrarily closely with positive probability: $\psi$ is in the
support if, for every $\nu>0$, a rule drawn from $\rho$ is within distance
$\nu$ of $\psi$ with positive probability,
\[
  \operatorname{supp}\rho=\Big\{\psi:\ \mathbb P_{\psi'\sim\rho}\big[d_0(\psi',\psi)<\nu\big]>0
  \ \text{for every }\nu>0\Big\}.
\]
A rule in $\operatorname{supp}\rho$ need not be drawn with positive probability. For
the switch sampler, any particular rule has probability zero, but rules
arbitrarily close to it are drawn with positive probability.

\subsection{Conventions}\label{sec:conventions}
 
Throughout the appendix we write $q$ for $q_\phi$. The distance $d$ on $\Theta$ is induced by a norm, $B(c,t)=\{\theta:d(\theta,c)\le t\}$ is the closed ball, and $\mathrm{vol}$ is the Lebesgue measure on $\mathbb R^{d_\theta}$. By \cref{ass:A}, $\Theta$ is open and, for every $y$, $p(\cdot\mid y)$ and $q(\cdot\mid y)$ have continuous densities that are strictly positive on $\Theta$. We recall the objects of the main text. For a reference point $c\in\Theta$, the radial mass function of the estimator is 
\[
  m_c(t\mid y)=\mathbb P_{\tilde\theta\sim q(\cdot\mid y)}\big[\tilde\theta\in B(c,t)\big],
  \qquad t\ge0 .
\]
For a rule $\psi$, the credibility value is $s_\psi(\theta,y)=m_{\psi(y)}\big(d(\theta,\psi(y))\mid y\big)$, the conditional coverage is $F_\psi(\alpha\mid y)=\mathbb P_{\theta\sim p(\cdot\mid y)}[s_\psi(\theta,y)\le\alpha]$, and the expected coverage is $F_\psi(\alpha)=\int F_\psi(\alpha\mid y)\,p(y)\,dy$. For a constant rule $\psi\equiv c$ we write $F_c(\alpha\mid y)$. Finally,
\[
  I_\psi=\int_0^1\big(F_\psi(\alpha)-\alpha\big)^2d\alpha,
  \qquad
  \mathcal D_\rho(p,q)=\Eb_{\psi\sim\rho}\big[I_\psi\big].
\]
Two remarks are used repeatedly. First, the conditional coverage at $y$ depends on the rule only through its reference point at $y$: $F_\psi(\alpha\mid y)=F_{\psi(y)}(\alpha\mid y)$. Second, $\alpha\mapsto F_\psi(\alpha\mid y)$ and $\alpha\mapsto F_\psi(\alpha)$ are cumulative distribution functions, hence non-decreasing and right-continuous.

\subsection{Preliminary facts}\label{sec:prelim}

We restate and prove \textbf{\cref{lem:post-calibrated}}.

\textit{Under \cref{ass:A}, if $q(\cdot\mid y)=p(\cdot\mid y)$ for every $y$, then $F_\psi(\alpha\mid y)=\alpha$ for every rule $\psi$, every level $\alpha\in[0,1]$ and every $y$, and hence $F_\psi(\alpha)=\alpha$.}

\begin{proof}
A correct estimator is calibrated, no matter how the regions are chosen. Let $R_\alpha$ be any credible region map of $q$. Factoring the joint density as $p(\theta,y)=p(y)\,p(\theta\mid y)$ and exchanging the order of integration, we obtain
\begin{align*}
  \mathrm{ECP}(q,\alpha,R_\alpha)
  &=\mathbb P_{(\theta,y)\sim p}\big[\theta\in R_\alpha(y)\big]
   =\iint \mathbf 1_{\theta\in R_\alpha(y)}\,p(\theta,y)\,d\theta\,dy\\
  &=\int p(y)\left(\int \mathbf 1_{\theta\in R_\alpha(y)}\,p(\theta\mid y)\,d\theta\right)dy
   =\int p(y)\left(\int_{R_\alpha(y)} q(\theta\mid y)\,d\theta\right)dy\\
  &=\int p(y)\,\alpha\,dy=\alpha ,
\end{align*}
where the fourth equality uses $p(\cdot\mid y)=q(\cdot\mid y)$, and the fifth uses the definition of a credible region. The inner integral equals $\alpha$ for every $y$, so the statement holds at each measurement, before averaging over $y$. For the TARP region of a rule $\psi$, the event $\theta\in R^{\mathrm{TARP}}_{\alpha,\psi}(y)$ is the event $s_\psi(\theta,y)\le\alpha$ (Section~\ref{sec:credibility}): the credibility value of $\theta$ is the mass of the ball where $\theta$ is at the border, so it is at most $\alpha$ exactly when $\theta$ lies in the ball of mass $\alpha$. Hence $F_\psi(\alpha\mid y)=\mathbb P_{\theta\sim p(\cdot\mid y)}\big[\theta\in R^{\mathrm{TARP}}_{\alpha,\psi}(y)\big]=\alpha$ for every $y$, and averaging over $p(y)$ gives $F_\psi(\alpha)=\alpha$.
\end{proof}
 
\begin{lemma}[Radial mass function]\label{lem:radial}
Under \cref{ass:A}, for every $y$:
\begin{itemize}
  \item[(i)] $(c,t)\mapsto m_c(t\mid y)$ is continuous on $\Theta\times[0,\infty)$;
  \item[(ii)] for every $c\in\Theta$ there is $t_c>0$ such that $t\mapsto m_c(t\mid y)$ is strictly increasing on $[0,t_c]$, with
        $m_c(0\mid y)=0$.
\end{itemize}
\end{lemma}
\begin{proof}
(i) Since $\Theta\times[0,\infty)$ is a metric space, it suffices to check continuity along sequences. Let $(c_n,t_n)_{n\in\mathbb N}$ be a sequence converging to $(c,t)$. For every $\theta$ with $d(\theta,c)\ne t$, the indicator $\mathbf 1_{d(\theta,c_n)\le t_n}$ converges to $\mathbf 1_{d(\theta,c)\le t}$. The sphere $\{\theta:d(\theta,c)=t\}$ has zero volume, hence zero mass under $q(\cdot\mid y)$, so the convergence holds $q$-almost everywhere and dominated convergence gives $m_{c_n}(t_n\mid y)\to m_c(t\mid y)$. (ii) Since $\Theta$ is open, there is $t_c>0$ with $B(c,t_c)\subseteq\Theta$. For $0\le t<t'\le t_c$, the difference $m_c(t'\mid y)-m_c(t\mid y)$ is the $q$-mass of the shell $B(c,t')\setminus B(c,t)$, which has positive volume and lies in $\Theta$, where $q(\cdot\mid y)>0$. Hence it is positive. Finally $m_c(0\mid y)$ is the mass of the point $c$, which is zero.
\end{proof}

\begin{proposition}[Half-spaces detect functions]\label{prop:cw}
Let $f$ be a bounded function on $\mathbb R^{d_y}$ such that
\[
  \int_H f(y)\,p(y)\,dy=0
  \quad\text{for every half-space } H=\{y:\langle u,y\rangle\le t\},
\]
with $u$ a unit vector and $t\in\mathbb R$. Then $f(y)=0$ for $p(y)$-almost every $y$.
\end{proposition}
\begin{proof}
The idea is that integrals over half-spaces determine all integrals of functions of a projection $\langle u,y\rangle$, and these in turn determine the Fourier transform of $f p$. Write $g(y)=f(y)\,p(y)$, which is integrable since $f$ is bounded, and fix a unit vector $u$.
 
\emph{Slabs.} Letting $t\to\infty$ in the hypothesis gives $\int g=0$. Subtracting the hypothesis at two thresholds $a<b$ gives $\int_{\{a<\langle u,y\rangle\le b\}}g(y)\,dy=0$. Hence $\int h(\langle u,y\rangle)\,g(y)\,dy=0$ for every step function $h$, that is, every finite combination of indicators of intervals $(a,b]$.
 
\emph{Trigonometric functions.} Fix $\lambda\in\mathbb R$, $\kappa>0$ and $n\ge1$. Let $h_n$ be the step function equal to $\cos(\lambda a)$ on each interval $(a,a+1/n]$ of a regular grid of $[-\kappa,\kappa]$, and to $0$ outside $[-\kappa,\kappa]$. Then $|\cos(\lambda s)-h_n(s)|\le|\lambda|/n$ for $|s|\le\kappa$, and $\le1$ otherwise, so 
\[
  \Big|\int\cos(\lambda\langle u,y\rangle)\,g(y)\,dy\Big|
  =\Big|\int\big(\cos(\lambda\langle u,y\rangle)-h_n(\langle u,y\rangle)\big)\,g(y)\,dy\Big|
  \le\sup|f|\Big(\frac{|\lambda|}{n}+\mathbb P_{y\sim p}\big[|\langle u,y\rangle|>\kappa\big]\Big).
\]
Letting $n\to\infty$ and then $\kappa\to\infty$ shows that the left-hand side is zero. The same argument applies to $\sin$, so $\int e^{i\lambda\langle u,y\rangle}g(y)\,dy=0$.
 
\emph{Conclusion.} Every vector $v\in\mathbb R^{d_y}$ is of the form $\lambda u$, so the Fourier transform $\hat g(v)=\int e^{i\langle v,y\rangle}g(y)\,dy$ vanishes everywhere. An integrable function whose Fourier transform vanishes is zero almost everywhere, so $f(y)\,p(y)=0$ for almost every $y$, that is, $f(y)=0$ for $p(y)$-almost every $y$. This is the Cram\'er--Wold argument
\citep{cramer1936some}.
\end{proof}
 
Half-spaces are used only through \cref{prop:cw}: any other family of sets for which this proposition holds can replace them in (R2).

\subsection{From zero loss to calibration on the support}\label{sec:step-support}
 
\begin{lemma}[Continuity]\label{lem:cont}
Under \cref{ass:A}, the map $\psi\mapsto I_\psi$ is continuous for the metric $d_0$.
\end{lemma}
\begin{proof}
Let $\psi_n\to\psi$ for $d_0$, that is, in $p(y)$-probability. It suffices to show that every subsequence has a further subsequence along which $I_{\psi_n}\to I_\psi$. Convergence in probability gives a subsequence, still denoted $\psi_n$, such that $\psi_n(y)\to\psi(y)$ for $p(y)$-almost every $y$. For such $y$ and every $\theta$,
\[
  s_{\psi_n}(\theta,y)=m_{\psi_n(y)}\big(d(\theta,\psi_n(y))\mid y\big)
  \longrightarrow
  m_{\psi(y)}\big(d(\theta,\psi(y))\mid y\big)=s_\psi(\theta,y),
\]
by continuity of $d$ and joint continuity of the radial mass function (\cref{lem:radial}). So $s_{\psi_n}\to s_\psi$ almost surely under $p(\theta,y)$, hence in distribution, and $F_{\psi_n}(\alpha)\to F_\psi(\alpha)$ at every continuity point of $F_\psi$, that is, for all but countably many $\alpha$. Since $|F_{\psi_n}(\alpha)-\alpha|\le1$, bounded convergence gives $I_{\psi_n}\to I_\psi$.
\end{proof}
 
\begin{lemma}\label{lem:support}
Under \cref{ass:A}, if $\mathcal D_\rho(p,q)=0$, then $F_\psi(\alpha)=\alpha$ for every $\psi\in\operatorname{supp}\rho$ and every $\alpha\in[0,1]$.
\end{lemma}
\begin{proof}
Since $I_\psi\ge0$ and $\Eb_{\psi\sim\rho}[I_\psi]=0$, we have $I_\psi=0$ for $\rho$-almost every $\psi$. The set $Z=\{\psi:I_\psi=0\}$ is closed by \cref{lem:cont}, and $\rho(Z)=1$. If some $\psi\in\operatorname{supp}\rho$ were outside $Z$, a small ball around $\psi$ would lie in the open complement of $Z$ and have positive probability under $\rho$, a contradiction. Hence $I_\psi=0$ on $\operatorname{supp}\rho$, so $F_\psi(\alpha)=\alpha$ for almost every $\alpha$, and for every $\alpha$ by right-continuity of both sides.
\end{proof}

\subsection{From calibration to invariance to the reference point}\label{sec:step-invariance}
 
Every rule in $\operatorname{supp}\rho$ is now calibrated, but calibration is an average over measurements, in which errors at different measurements can cancel. This step obtains a statement at each measurement: the coverage curve at $y$ does not depend on the reference point. As in the main-text example, it compares a rule $\psi'$ with the rule patched from $\psi$ on a half-space, which isolates what happens on that half-space.
 
\begin{lemma}[Invariance to the reference point]\label{lem:loc}
Let $\rho$ be admissible with family $\Psi$, and suppose that every rule in $\operatorname{supp}\rho$ is calibrated. Then, for $p(y)$-almost every $y$, the coverage curve at $y$ does not depend on the rule of $\Psi$:
\[
  F_\psi(\alpha\mid y)=F_{\psi'}(\alpha\mid y)
  \qquad\text{for all }\psi,\psi'\in\Psi\text{ and all }\alpha\in[0,1].
\]
\end{lemma}
 
\begin{proof}
\emph{Step 1: coverage of a patched rule.} Fix $\psi,\psi'\in\Psi$, a level $\alpha$, and a half-space $H$. The coverage at a measurement $y$ depends on the rule only through its reference point at $y$. The patched rule $\psi\oplus_H\psi'$ has reference point $\psi(y)$ for $y\in H$ and $\psi'(y)$ for $y\notin H$, so its coverage curve at $y$ is $F_\psi(\alpha\mid y)$ on $H$ and $F_{\psi'}(\alpha\mid y)$ outside $H$. Averaging over measurements,
\[
  F_{\psi\oplus_H\psi'}(\alpha)
  =\int_H F_\psi(\alpha\mid y)\,p(y)\,dy+\int_{y\notin H}F_{\psi'}(\alpha\mid y)\,p(y)\,dy .
\]
Subtracting $F_{\psi'}(\alpha)=\int F_{\psi'}(\alpha\mid y)\,p(y)\,dy$, the part outside $H$ cancels and
\[
  F_{\psi\oplus_H\psi'}(\alpha)-F_{\psi'}(\alpha)=\int_H\Delta(y)\,p(y)\,dy,
  \qquad
  \Delta(y)=F_\psi(\alpha\mid y)-F_{\psi'}(\alpha\mid y).
\]
Here $\Delta(y)$ is the difference of coverage at $y$ between the two reference points $\psi(y)$ and $\psi'(y)$, and $|\Delta|\le1$.
 
\emph{Step 2: both rules are calibrated.} The rule $\psi'$ is in $\operatorname{supp}\rho$ because $\Psi\subseteq\operatorname{supp}\rho$, and the patched rule $\psi\oplus_H\psi'$ is in $\operatorname{supp}\rho$ by (R2). Both are calibrated, so both coverages equal $\alpha$ and the left-hand side of Step 1 is zero:
\[
  \int_H\Delta(y)\,p(y)\,dy=0\qquad\text{for every half-space }H.
\]
 
\emph{Step 3: from half-spaces to each measurement.} By \cref{prop:cw}, applied to
$f=\Delta$, we get $\Delta(y)=0$ for $p(y)$-almost every $y$. In other words, $F_\psi(\alpha\mid y)=F_{\psi'}(\alpha\mid y)$ for every measurement $y$ except those in a set $N_{\psi,\psi',\alpha}$ that has probability zero under $p(y)$.
 
\emph{Step 4: all pairs and all levels at once.} The exceptional set of Step 3 depends on $\psi$, $\psi'$ and $\alpha$. Let $N$ be the union of these sets over all pairs $\psi,\psi'\in\Psi$ and all \textit{rational} levels $\alpha\in[0,1] \cap \mathbb{Q}$. There are countably many of them, so $N$ still has probability zero. For $y\notin N$, the two coverage curves agree at every rational level. Both are distribution functions, hence right-continuous, and every $\alpha\in[0,1)$ is a limit of rationals from the right, so they agree at every level; at $\alpha=1$ both equal one.
\end{proof}

\subsection{From invariance to the reference point to equal densities}\label{sec:step-densities}
 
\begin{lemma}\label{lem:balls}
Fix $y$ such that the reference points $\{\psi(y):\psi\in\Psi\}$ are dense in $\Theta$ and $F_\psi(\cdot\mid y)$ is the same function $G_y$ for all $\psi\in\Psi$. Then $p(\cdot\mid y)=q(\cdot\mid y)$.
\end{lemma}
\begin{proof}
Let $c=\psi(y)$ for some $\psi\in\Psi$, write $m=m_c(\cdot\mid y)$, and take $t_c$ from \cref{lem:radial}. For $0<t<t_c$, $m$ is strictly increasing on $[0,t_c]$ and non-decreasing beyond, so $m(D)\le m(t)$ if and only if $D\le t$. Applying this to $D=d(\theta,c)$ with $\theta\sim p(\cdot\mid y)$,
\[
  G_y\big(m(t)\big)=F_c\big(m(t)\mid y\big)
  =\mathbb P_{\theta\sim p(\cdot\mid y)}\big[\theta\in B(c,t)\big].
\]
Since $m(t)=\mathbb P_{\theta\sim q(\cdot\mid y)}[\theta\in B(c,t)]$, dividing gives
\[
  \frac{G_y(\alpha)}{\alpha}
  =\frac{\mathbb P_{\theta\sim p(\cdot\mid y)}\big[\theta\in B(c,t)\big]}
         {\mathbb P_{\theta\sim q(\cdot\mid y)}\big[\theta\in B(c,t)\big]},
  \qquad \alpha=m(t).
\]
As $t\downarrow0$, both densities are continuous and positive at $c$, so each probability equals the density at $c$ times $\mathrm{vol}(B(c,t))$, up to a factor $1+o(1)$. The volumes cancel and the right-hand side tends to $p(c\mid y)/q(c\mid y)$. On the left, $m$ is a continuous increasing bijection from $(0,t_c)$ onto $(0,m(t_c))$, so $\alpha\downarrow0$ exactly when $t\downarrow0$. Therefore $L=\lim_{\alpha\downarrow0}G_y(\alpha)/\alpha$ exists and
\[
  p(c\mid y)=L\,q(c\mid y).
\]
The limit $L$ depends on $G_y$ only, not on $c$, so this identity holds at every reference point $c=\psi(y)$, $\psi\in\Psi$, a dense subset of $\Theta$. Both densities are continuous, so it holds on all of $\Theta$. Integrating over $\Theta$ gives $L=1$, hence $p(\cdot\mid y)=q(\cdot\mid y)$.
\end{proof}
 
\subsection{Proof of Theorem~\ref{thm:proper}}\label{sec:step-theorem}

\textit{Under \cref{ass:A} and for any admissible $\rho$,
\[
  \mathcal D_\rho(p,q)=0\iff q(\cdot\mid y)=p(\cdot\mid y)\ \text{for $p(y)$-almost every }y.
\]}
\begin{proof}
($\Leftarrow$) By \cref{lem:post-calibrated}, $F_\psi(\alpha)=\alpha$ for every rule
$\psi$ and every $\alpha$, so $I_\psi=0$ for every $\psi$ and $\mathcal D_\rho=0$. This
direction holds for any $\rho$.
 
($\Rightarrow$) Suppose $\mathcal D_\rho=0$. By \cref{lem:support}, every rule in
$\operatorname{supp}\rho$ is calibrated. By \cref{lem:loc}, there is a null set $N_2$
outside which $F_\psi(\cdot\mid y)$ is the same function for all $\psi\in\Psi$. By
(R1), there is a null set $N_1$ outside which the reference points $\{\psi(y):\psi\in\Psi\}$
are dense in $\Theta$. For every $y\notin N_1\cup N_2$, \cref{lem:balls} applies and
gives $p(\cdot\mid y)=q(\cdot\mid y)$. Since $N_1\cup N_2$ is null, this holds for
$p(y)$-almost every $y$.
\end{proof}

\section{Training algorithm}\label{sec:appendix-algo}

\cref{alg:tcvm} summarizes one training step with the \tcvm regularizer. The base loss $\mathcal L_{\mathrm{base}}$ is the training objective of the generative model, such as denoising score matching for diffusion models or the negative log-likelihood for normalizing flows.

\paragraph{Complexity} For each rule, computing the $N$ credibility values requires $NM$ distance evaluations, and the closed-form statistic of \autoref{eq:cvm} only requires sorting them, at cost $\mathcal O(N\log N)$. Its gradient follows analytically from the ranks, so no differentiable sorting operator is needed. One step therefore costs $\mathcal O\big(R\,N\,(M+\log N)\big)$ operations on top of drawing the $NM$ posterior samples. In practice, the cost is dominated by sampling. For a diffusion model with a $T$-step sampler, a step requires $NMT$ network evaluations, and backpropagating through the chains that carry gradient multiplies their cost, and the memory they require, by a factor proportional to $T$. Training cost thus grows linearly with the sampler depth, which is why we train with $T=20$ (\cref{sec:ip}, Appendix~\ref{sec:appendix-ip}).

\section{Unbiased estimator of the Cram\'{e}r--von Mises statistic}\label{sec:app-unbiased-estimator}

For a fixed rule $\psi$, the inner integral of \autoref{eq:loss} is
\[
I_\psi=\int_0^1\big(F_\psi(\alpha)-\alpha\big)^2\,d\alpha .
\]
Given the credibility values $s_1,\dots,s_N$ of a batch, \autoref{eq:cvm} replaces $F_\psi$ by the empirical CDF $\hat F_\psi(\alpha)=\frac1N\sum_i\mathbf 1_{s_i\le\alpha}$, so that $\cvm(s_{1:N})=\int_0^1\big(\hat F_\psi(\alpha)-\alpha\big)^2\,d\alpha$. Writing $g_i(\alpha)=\mathbf 1_{s_i\le\alpha}-\alpha$, we have $\hat F_\psi(\alpha)-\alpha=\frac1N\sum_i g_i(\alpha)$, and
\[
\cvm(s_{1:N})=\frac1{N^2}\sum_{i=1}^N\sum_{j=1}^N h(s_i,s_j),
\qquad
h(s,s')=\int_0^1\big(\mathbf 1_{s\le\alpha}-\alpha\big)\big(\mathbf 1_{s'\le\alpha}-\alpha\big)\,d\alpha .
\]
This is an average over all $N^2$ pairs $(i,j)$, including the $N$ diagonal pairs $i=j$, that is, a V-statistic \citep{mises1947asymptotic}. The U-statistic \citep{Hoeffding} $\cvm_U$ of \autoref{eq:cvm-u} averages $h$ over the $N(N-1)$ pairs with $i\neq j$ only.

\begin{proposition}[Bias of the plug-in estimator]\label{prop:cvm-u}
Let $s_1,\dots,s_N$, with $N\ge2$, be independent with common CDF $F_\psi$ on $[0,1]$. Then
\[
\Eb\big[\cvm_U(s_{1:N})\big]=I_\psi,
\qquad
\Eb\big[\cvm(s_{1:N})\big]=I_\psi+\frac1N\int_0^1F_\psi(\alpha)\big(1-F_\psi(\alpha)\big)\,d\alpha .
\]
\end{proposition}

\begin{proof}
Fix a level $\alpha$. For $i\neq j$, the values $s_i$ and $s_j$ are independent, so
\[
\Eb\big[g_i(\alpha)\,g_j(\alpha)\big]=\Eb\big[g_i(\alpha)\big]\,\Eb\big[g_j(\alpha)\big]=\big(F_\psi(\alpha)-\alpha\big)^2,
\]
which is the integrand of $I_\psi$. For $i=j$, we obtain the second moment instead,
\[
\Eb\big[g_i(\alpha)^2\big]=\mathrm{Var}\big(\mathbf 1_{s_i\le\alpha}\big)+\big(F_\psi(\alpha)-\alpha\big)^2
=F_\psi(\alpha)\big(1-F_\psi(\alpha)\big)+\big(F_\psi(\alpha)-\alpha\big)^2 .
\]
By Fubini's theorem, $\Eb[h(s_i,s_j)]$ is the integral over $\alpha$ of these expectations. Averaging over the $N(N-1)$ off-diagonal pairs gives $\Eb[\cvm_U]=I_\psi$, and averaging over all $N^2$ pairs, of which $N$ are diagonal, gives the expectation of $\cvm$.

It remains to check that \autoref{eq:cvm-u} is the off-diagonal average of $h$. Using $\int_0^1\mathbf 1_{\max(s,s')\le\alpha}\,d\alpha=1-\max(s,s')$, $\int_0^1\alpha\,\mathbf 1_{s\le\alpha}\,d\alpha=(1-s^2)/2$ and $\int_0^1\alpha^2\,d\alpha=1/3$,
\[
h(s,s')=\tfrac12\big(s^2+s'^2\big)-\max(s,s')+\tfrac13 .
\]
Summing over the ordered pairs $i\neq j$, the first term sums  to $(N-1)\sum_i s_i^2$, and the $k$-th smallest value $s_{(k)}$ is the maximum of the $2(k-1)$ ordered pairs it forms with the smaller values. Dividing by $N(N-1)$ gives \autoref{eq:cvm-u}.
\end{proof}

Note that the bias term does not vanish at the posterior. When $F_\psi(\alpha)=\alpha$, it equals $\frac1N\int_0^1\alpha(1-\alpha)\,d\alpha=\frac1{6N}$. It also shifts the minimizer of the expected loss.

\section{Scoring rules and loss landscapes}\label{sec:appendix-scoring}

\subsection{Definitions}

The continuous ranked probability score (CRPS) compares the cumulative distribution function (CDF) of the model with the CDF of a point mass at the observation. Part of its appeal is that CDFs are easy to estimate from an ensemble of generated samples. For models that are easy to sample from but have no closed-form CDF, the CRPS admits the energy form
\[
\mathrm{CRPS}(\qy, \theta) = 2\, \mathbb E_{\hat \theta \sim \qy} |\hat \theta - \theta| - \mathbb E_{\hat \theta, \hat \theta' \sim \qy} |\hat \theta- \hat \theta'| .
\]
The first term penalizes a model whose samples are far from the true $\theta$ on average, and the second rewards samples that are spread out, so that a model cannot minimize the CRPS by collapsing to a point prediction.

Unlike CDFs, this energy form generalizes directly to several dimensions by replacing the absolute value with the Euclidean norm, which gives the energy score of \cref{sec:related}. On the class of distributions with finite first moment, the energy score is strictly proper:
\[
\mathbb E_{\theta \sim p(\cdot \mid y)} [\mathrm{ES}(\qy, \theta)] \geq \mathbb E_{\theta \sim p(\cdot \mid y)} [\mathrm{ES}(p(\cdot\mid y), \theta)],
\]
with equality if and only if $\qy = p(\cdot \mid y)$. Like CRPS, it can be evaluated from samples of a generative model alone.

In high dimension, the CRPS is often computed for each coordinate separately rather than replaced by the energy score. We call this the pixel-wise CRPS,
\[
\mathrm{pCRPS}(\qy, \theta) = \sum_{j=1}^{d_\theta} \mathrm{CRPS}\big(\qy^{(j)}, \theta^{(j)}\big),
\]
where the superscript $j$ denotes restriction to the $j$-th coordinate. Its motivation is often computational, since it parallelizes across pixels. It is however minimized by any model with the correct one-dimensional marginals, and therefore does not constrain the dependence between coordinates.

\subsection{Strict propriety and training}

Strict propriety is a statement about the global minimizer of the population loss, and two gaps separate it from what training achieves. First, the loss is minimized over a finite training set, and the minimizer of this empirical approximation need not be the minimizer of the population loss. Second, gradient descent generally finds local rather than global minima. Two strictly proper scoring rules share the same global minimizer, but they can have very different loss landscapes, with different local minima, so that training with one can yield much better coverage than training with the other. \cref{sec:toy_model} illustrates this on a Gaussian model.

\subsection{Loss landscapes}\label{sec:toy_model}

We examine the same setting as the inverse problem in \cref{fig:fig1}, setting $\theta$ to have dimension 8 and $Y$ dimension 4. 

\begin{figure}[h]
    \centering
    \includegraphics[width=1.0\linewidth]{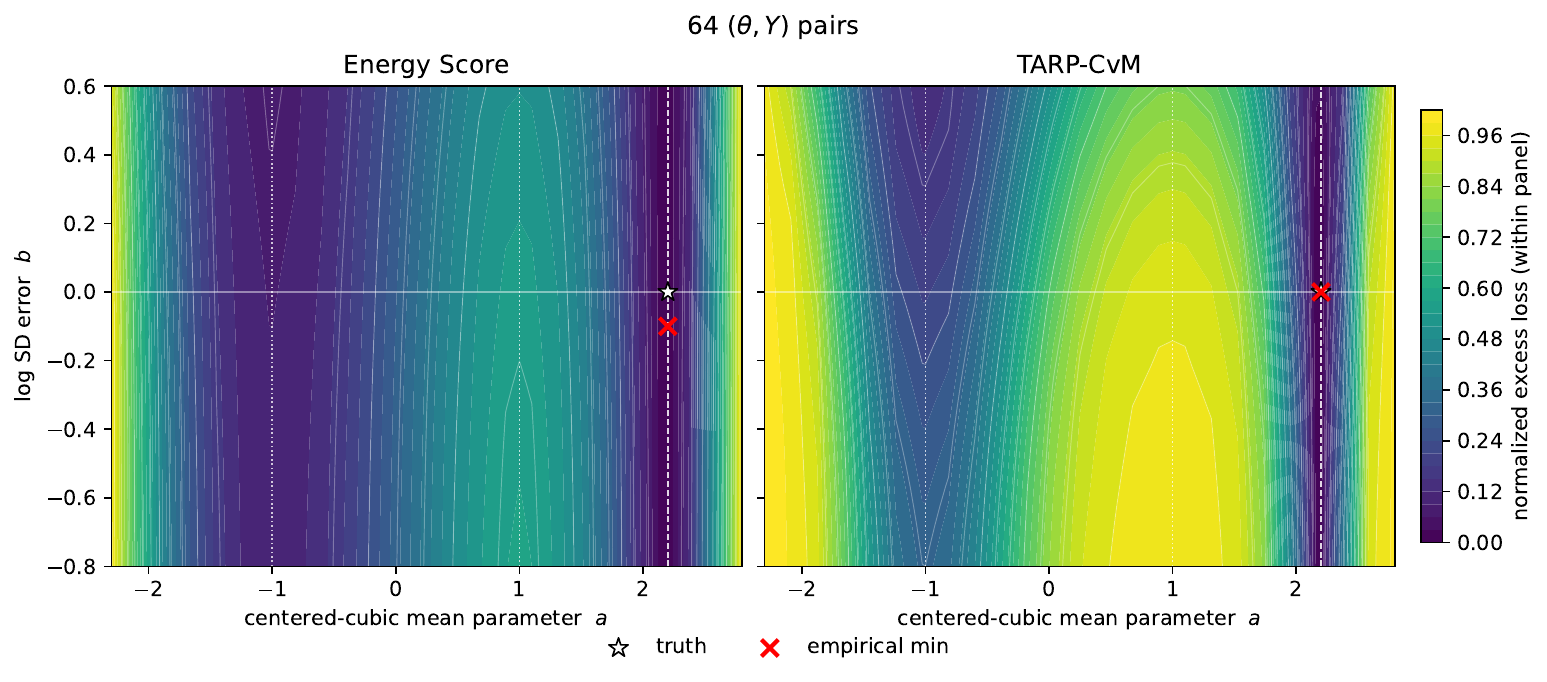}
    \caption{Loss landscapes for the toy model described in \cref{sec:toy_model}.}
    \label{fig:loss_landscapes}
\end{figure}

We identify the true parameters of the posterior distribution $\theta\mid Y$ and perturb them by deviating the mean in the direction of largest variance and the scale of the variance in that same direction. See the equation below for this perturbation model:
\begin{equation*}
    q_{a, b}(\theta \mid y)=\operatorname{Normal}\left(W^* y+h(a) \sqrt{  \lambda_j} v_j, V \operatorname{diag}\left(\lambda_1, \ldots, e^{2 b} \lambda_j, \ldots, \lambda_D\right) V^{\top}\right),
\end{equation*}
where $a,b$ are the mean and scale perturbation parameters, respectively. $h(a) = a^3 -3a -4.048$ parameterizes the mean perturbation and vanishes at $a=2.2$. The scale perturbation vanishes at $b=0$. We see in \cref{fig:loss_landscapes} the loss landscapes for Energy Score and \tcvm in terms of $a$ and $b$.

\begin{figure}[h]
    \centering
    \includegraphics[width=1.0\linewidth]{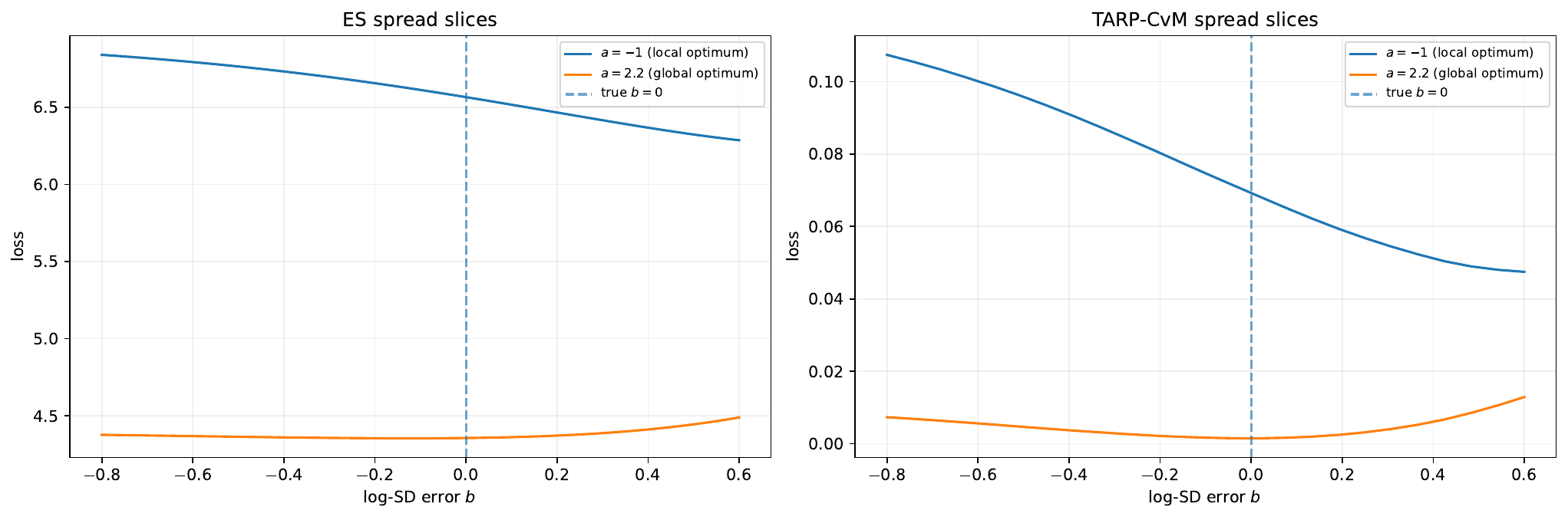}
    \caption{A slice of the loss landscape described in \cref{sec:toy_model}.}
    \label{fig:loss_slice}
\end{figure}

In \cref{fig:loss_slice}, we have slices of the loss functions at the local and global minima of $a$, the mean parameter. We see that \tcvm has much more sensitivity to changing the spread parameter, $b$, when $a$ is at the local minimum than energy score does. In other words, we are comparing the steepness of the blue lines in each plot.

Intuitively, when one gets stuck in a local minimum for the mean parameter, \tcvm recovers correct coverage by pushing the conditional distribution toward wider spread to compensate for the bias in the mean. In contrast, the energy score’s loss function is relatively flat with respect to the spread parameter in this local optimum. So, when noisy gradients provide imperfect optimization of this population loss, the energy score is less likely to correct coverage when gradient descent is near a local minimum for the conditional mean, since the signal in the loss is too flat to overcome gradient noise.
 
 \section{Experiments}

\subsection{Toy model}\label{sec:appendix-toy}

\paragraph{Setup} The parameters have a Gaussian prior $\theta\sim\mathcal N(0,\Sigma_0)$, and the measurements follow $y=A\theta+\varepsilon$, with $\varepsilon\sim\mathcal N(0,\sigma^2 I)$, $d_y=4$, and $d_\theta=8$, so that $A$ cannot be inverted. The posterior is Gaussian,
\[
  p(\theta\mid y)=\mathcal N(W^*y,\,S^*),\qquad
  W^*=\Sigma_0A^\top\Sigma_y^{-1},\qquad
  S^*=\Sigma_0-W^*A\Sigma_0,
\]
where $\Sigma_y=A\Sigma_0A^\top+\sigma^2I$ is the covariance of the measurements. The estimator is $q_\phi(\theta\mid y)=\mathcal N(Wy,S)$, where $\phi=(W, S)$ are parameters to fit.

\paragraph{HPD calibration does not identify the posterior} The density of $\qp(\cdot\mid y)$ is a decreasing function of the Mahalanobis norm $m(\theta)=(\theta-Wy)^\top S^{-1}(\theta-Wy)$, so its HPD regions are the ellipsoids $\{\theta: m(\theta)\le t_\alpha\}$ with mass $\alpha$ under $\qp(\cdot\mid y)$. Under $p(\theta,y)$, the residual $e=\theta-Wy$ decomposes as $(\theta-W^*y)+(W^*-W)y$. The first term has distribution $\mathcal N(0,S^*)$ given $y$, hence is independent of $y$, and the second has covariance $(W^*-W)\Sigma_y(W^*-W)^\top$, so that marginally
\[
  e\sim\mathcal N(0,C_W),\qquad C_W=S^*+(W^*-W)\,\Sigma_y\,(W^*-W)^\top .
\]
If $S=C_W$, the residual of the true parameter has the same distribution as $\tilde\theta-Wy$ for a draw $\tilde\theta\sim\qp(\cdot\mid y)$. The true parameter then falls in each HPD region with probability $\alpha$, so the estimator is HPD-calibrated, for every $W$.

\paragraph{Points of the manifold} At $W=W^*$, $S=S^*$ is the true posterior. At $W=0$, $S=S^*+W^*\Sigma_yW^{*\top}=S^*+W^*A\Sigma_0=\Sigma_0$, so the prior is HPD-calibrated. For every $W\neq W^*$, the mean $Wy$ is biased and $S-S^*=(W^*-W)\Sigma_y(W^*-W)^\top$ is positive semi-definite and nonzero, so the estimator is wider than the posterior at every measurement: marginal calibration is reached by trading the bias of the mean for extra spread.
 
 \subsection{Simulation Based Inference}\label{sec:appendix-sbi}

 \paragraph{Problems.} We use the simulators of \citet{hermans2022a} with the problem settings of \citet{falkiewicz2023}. For each problem we generate up to $2^{16}$ simulations for training, with $10^4$ for validation, and $10^4$ held-out pairs for evaluation.
 \begin{itemize}
     \item \emph{SLCP} (simple likelihood, complex posterior) is used in the marginal variant of \citet{hermans2022a}, in which only two parameters are inferred and the likelihood becomes intractable. The parameters are $\theta\in[-3,3]^2$. The observation $y\in\mathbb R^8$ consists of four draws from a bivariate Gaussian with fixed mean $(0.7,-2.9)$, standard deviations $\theta_1^2$ and $\theta_2^2$, and correlation $\tanh(\rho)$, where nuisance parameter $\rho\sim\mathcal U(-3,3)$ is marginalized out. Since $\theta$ enters only through its squares, the posterior has four symmetric modes.
     
    \item \emph{Weinberg} infers the Fermi constant, normalized to its nominal value, $\theta\in[0.25,2]$, from $y\in[-1,1]^{20}$. The observation contains the cosines of the scattering angle of  twenty $e^+e^-\to\mu^+\mu^-$ events drawn by rejection sampling from the differential cross-section at a beam energy of $40$\,GeV.                                             
    \item \emph{Spatial SIR} infers the infection and recovery rates $\theta=(\beta,\gamma)\in[0,1]^2$ of a stochastic epidemic on a $50\times50$ lattice. The epidemic is seeded with $1+\mathrm{Poisson}(3)$ infected cells at random positions and propagated through 3\,$\times$\,3 neighborhoods. The observation $y\in\{0,1\}^{3\times50\times50}$ records the susceptible, infected and recovered state of every cell at time $t=1$.                      
    \item \emph{Lotka--Volterra} is a predator-prey Markov jump process started from $50$ predators and $100$ prey. The parameters $\theta\in[-4,1]^2$ are the log-rates of the two predator reactions. The two prey log-rates are drawn from the same prior and marginalized out. The observation $y\in\mathbb R^{2\times1001}$ is the trajectory of both populations.         
  \end{itemize}

 \paragraph{Architecture and training.} Neural posterior estimation \citep[NPE,][]{papamakarios2016fast, lueckmann2017flexible, deistler2022truncated, pmlr-v97-greenberg19a} fits a conditional density estimator $\qp(\theta\mid y)$ by maximum likelihood on simulated pairs, minimizing $-(1/N)\sum_i\log\qp(\theta_i\mid  y_i)$. This is equivalent to minimizing the forward KL divergence to the posterior, averaged over $p(y)$. We use the architecture of \citet{falkiewicz2023}, implemented with the \textsc{sbi} library \citep{tejero-cantero2020sbi}. It is a conditional neural spline flow \citep{durkan2019neural}, with rational-quadratic spline transforms (10 bins, hidden width 64) and a Gaussian base distribution. There are three transforms, each a coupling layer followed by a linear layer, except for the one-dimensional Weinberg problem, which uses a single transform whose spline parameters depend on the conditioning only. The flow is conditioned on an embedding of $y$ given by an MLP with three hidden layers of width 64, SELU activations and a 10-dimensional output. The 7500-dimensional Spatial SIR observations and the 2002-dimensional Lotka-Volterra trajectories are flattened before embedding. 
 
 Parameters and observations are standardized per coordinate using the training data. We train with Adam, using a learning rate of $10^{-3}$, a batch size of 128, gradient-norm clipping at 5, for 50 epochs without early stopping. We keep the checkpoint with the lowest validation loss on the $10^4$ validation simulations. All three objectives of Table~\ref{tab:sbi} share this training recipe. 
 
 For CalNPE, we use the conservativeness objective of \citet{falkiewicz2023}, which penalizes only coverage below the nominal level, with weight $\lambda=100$, the default of their released code, instead of the published $\lambda=5$, which gave worse results in our experiments. HPD credibilities are estimated from the density of the estimator at 16 draws from the prior, with straight-through gradients for the indicators and differentiable sorting \citep{blondel2020fast}.
 
 \tcvm uses weight $\lambda=30$ and $M=16$ reparameterized posterior draws per pair, so that both regularizers use $16$ flow evaluations per pair: CalNPE evaluates the density at $16$ prior draws, \tcvm draws $16$ posterior samples. Each step uses $R=48$ references in three groups of $16$, described below. The indicators are relaxed with a sigmoid at temperature $\tau=0.05$ times the median sample-to-reference distance of each reference, treated as a constant, and the loss is the closed-form CvM statistic of \autoref{eq:cvm}, computed per reference and averaged over the $R$ references.

\paragraph{Reference points without a forward operator} Without access to the forward operator to estimate $b(y)$, we resort to regressing the posterior mean. The point estimate $b(y)$ is an MLP (two hidden layers of width 128, SELU) trained beforehand on the flow's own training pairs with a mean-squared-error loss, with AdamW (learning rate $10^{-3}$, weight decay $10^{-4}$, batch 256). The $R=48$ references of a step form three groups of $16$: (1) switch-rule references, (2) points drawn from the prior distribution and shared by the batch, and (3) the posterior draws of each pair, detached from the computation graph. The credibility of a posterior reference is computed from the $15$ other draws, so that $s_i$ remains uniform when $\qp=p$, by exchangeability.

 \begin{table}[ht]
\centering
\caption{SBI benchmark results for NPE, CalNPE \citep[$\lambda=100$,][]{falkiewicz2023} and \tcvm (Ours, $\lambda=30$, $M=16$), all trained with the same protocol, for simulation budgets $n \in \{2^{10}, 2^{13}, 2^{16}\}$. Mean $\pm$ standard deviation over five independently trained estimators per cell. Bold marks the best mean when it is significantly better than both other methods. HPD and TARP errors are the $L^1$ coverage errors under each credible-region map, $\mathbb{E}[\log q_\phi]$ is the expected log posterior density of the true parameters, MMD $z$ the z-score of an MMD test on the pairs $(\theta, b(y))$, and $|\mathrm{Mira} - 2/3|$ the deviation of the Mira score from its calibrated value.}
\label{tab:sbi}
\resizebox{\linewidth}{!}{%
\begin{tabular}{ll|ccc|ccc|ccc}
\toprule
 & & \multicolumn{3}{c}{HPD error $\downarrow$} & \multicolumn{3}{c}{TARP error, $y$-indep.\ $\downarrow$} & \multicolumn{3}{c}{TARP error, switch $\downarrow$} \\
\cmidrule(lr){3-5}\cmidrule(lr){6-8}\cmidrule(lr){9-11}
Problem & $n$ & NPE & CalNPE & Ours & NPE & CalNPE & Ours & NPE & CalNPE & Ours \\
\midrule
SLCP & $2^{10}$ & .096$_{\pm .017}$ & .086$_{\pm .014}$ & .101$_{\pm .018}$ & .015$_{\pm .004}$ & .012$_{\pm .002}$ & .015$_{\pm .005}$ & .013$_{\pm .003}$ & .012$_{\pm .002}$ & .014$_{\pm .004}$ \\
 & $2^{13}$ & .047$_{\pm .013}$ & .029$_{\pm .007}$ & .038$_{\pm .013}$ & .010$_{\pm .003}$ & .010$_{\pm .004}$ & .013$_{\pm .004}$ & .009$_{\pm .003}$ & .009$_{\pm .003}$ & .013$_{\pm .004}$ \\
 & $2^{16}$ & .013$_{\pm .007}$ & .006$_{\pm .004}$ & .018$_{\pm .007}$ & .007$_{\pm .002}$ & .008$_{\pm .002}$ & .006$_{\pm .002}$ & .007$_{\pm .002}$ & .008$_{\pm .002}$ & .006$_{\pm .001}$ \\
\addlinespace[2pt]
Weinberg & $2^{10}$ & .021$_{\pm .005}$ & \textbf{.010$_{\pm .004}$} & .031$_{\pm .011}$ & .011$_{\pm .003}$ & .010$_{\pm .003}$ & .014$_{\pm .004}$ & .047$_{\pm .003}$ & .050$_{\pm .004}$ & \textbf{.041$_{\pm .004}$} \\
 & $2^{13}$ & .034$_{\pm .005}$ & \textbf{.007$_{\pm .005}$} & .031$_{\pm .007}$ & .012$_{\pm .002}$ & .012$_{\pm .004}$ & .013$_{\pm .003}$ & .015$_{\pm .002}$ & .019$_{\pm .004}$ & .016$_{\pm .003}$ \\
 & $2^{16}$ & .016$_{\pm .006}$ & .008$_{\pm .003}$ & .013$_{\pm .005}$ & .009$_{\pm .003}$ & .008$_{\pm .003}$ & .009$_{\pm .002}$ & .009$_{\pm .003}$ & .011$_{\pm .002}$ & .008$_{\pm .002}$ \\
\addlinespace[2pt]
Spatial SIR & $2^{10}$ & .081$_{\pm .024}$ & \textbf{.045$_{\pm .021}$} & .093$_{\pm .012}$ & .029$_{\pm .008}$ & .027$_{\pm .006}$ & .034$_{\pm .003}$ & .041$_{\pm .010}$ & .041$_{\pm .016}$ & .027$_{\pm .003}$ \\
 & $2^{13}$ & .053$_{\pm .021}$ & \textbf{.017$_{\pm .006}$} & .041$_{\pm .016}$ & .025$_{\pm .006}$ & .020$_{\pm .006}$ & .027$_{\pm .004}$ & .021$_{\pm .004}$ & .022$_{\pm .005}$ & .021$_{\pm .005}$ \\
 & $2^{16}$ & .031$_{\pm .009}$ & .026$_{\pm .014}$ & .031$_{\pm .013}$ & .020$_{\pm .007}$ & .011$_{\pm .003}$ & .016$_{\pm .006}$ & .015$_{\pm .003}$ & .011$_{\pm .003}$ & .015$_{\pm .005}$ \\
\addlinespace[2pt]
Lotka--Volterra & $2^{10}$ & .082$_{\pm .016}$ & .064$_{\pm .008}$ & .081$_{\pm .008}$ & .016$_{\pm .003}$ & .015$_{\pm .005}$ & .013$_{\pm .006}$ & .067$_{\pm .004}$ & .067$_{\pm .005}$ & .067$_{\pm .003}$ \\
 & $2^{13}$ & .024$_{\pm .014}$ & .014$_{\pm .006}$ & .015$_{\pm .008}$ & .017$_{\pm .005}$ & .011$_{\pm .005}$ & .012$_{\pm .005}$ & .030$_{\pm .005}$ & .032$_{\pm .006}$ & .033$_{\pm .004}$ \\
 & $2^{16}$ & .014$_{\pm .007}$ & .010$_{\pm .004}$ & .013$_{\pm .010}$ & .014$_{\pm .008}$ & .014$_{\pm .005}$ & .012$_{\pm .004}$ & .014$_{\pm .007}$ & .017$_{\pm .002}$ & .013$_{\pm .003}$ \\
\bottomrule
\end{tabular}}

\vspace{6pt}

\resizebox{\linewidth}{!}{%
\begin{tabular}{ll|ccc|ccc|ccc}
\toprule
 & & \multicolumn{3}{c}{$\mathbb{E}[\log q_\phi]$ $\uparrow$} & \multicolumn{3}{c}{MMD $z$, $(\theta, b(y))$ $\downarrow$} & \multicolumn{3}{c}{$|\mathrm{MIRA} - \tfrac{2}{3}|$ $\downarrow$} \\
\cmidrule(lr){3-5}\cmidrule(lr){6-8}\cmidrule(lr){9-11}
Problem & $n$ & NPE & CalNPE & Ours & NPE & CalNPE & Ours & NPE & CalNPE & Ours \\
\midrule
SLCP & $2^{10}$ & -2.34$_{\pm 0.05}$ & -2.32$_{\pm 0.08}$ & -2.32$_{\pm 0.03}$ & 11.1$_{\pm 4.1}$ & 12.2$_{\pm 3.8}$ & 14.0$_{\pm 6.2}$ & .0017$_{\pm .0009}$ & .0033$_{\pm .0010}$ & .0022$_{\pm .0013}$ \\
 & $2^{13}$ & -1.41$_{\pm 0.08}$ & -1.38$_{\pm 0.02}$ & -1.35$_{\pm 0.06}$ & 5.5$_{\pm 2.7}$ & 6.3$_{\pm 3.0}$ & 5.6$_{\pm 2.4}$ & .0034$_{\pm .0010}$ & .0040$_{\pm .0014}$ & .0027$_{\pm .0016}$ \\
 & $2^{16}$ & -0.87$_{\pm 0.03}$ & -0.88$_{\pm 0.01}$ & -0.86$_{\pm 0.02}$ & 1.3$_{\pm 1.0}$ & 1.9$_{\pm 0.6}$ & 2.1$_{\pm 1.1}$ & .0045$_{\pm .0004}$ & .0051$_{\pm .0008}$ & .0048$_{\pm .0006}$ \\
\addlinespace[2pt]
Weinberg & $2^{10}$ & -0.11$_{\pm 0.01}$ & -0.10$_{\pm 0.01}$ & -0.11$_{\pm 0.01}$ & 56.3$_{\pm 6.2}$ & 59.3$_{\pm 13.6}$ & 58.2$_{\pm 8.9}$ & .0050$_{\pm .0017}$ & .0068$_{\pm .0015}$ & .0028$_{\pm .0015}$ \\
 & $2^{13}$ & 0.02$_{\pm 0.01}$ & 0.02$_{\pm 0.01}$ & 0.01$_{\pm 0.01}$ & 8.0$_{\pm 4.8}$ & 14.9$_{\pm 9.1}$ & 8.1$_{\pm 4.3}$ & .0014$_{\pm .0008}$ & .0039$_{\pm .0010}$ & .0013$_{\pm .0008}$ \\
 & $2^{16}$ & 0.10$_{\pm 0.00}$ & 0.09$_{\pm 0.01}$ & 0.09$_{\pm 0.01}$ & 3.8$_{\pm 3.1}$ & 7.1$_{\pm 5.3}$ & 5.3$_{\pm 3.4}$ & .0034$_{\pm .0012}$ & .0071$_{\pm .0009}$ & .0043$_{\pm .0006}$ \\
\addlinespace[2pt]
Spatial SIR & $2^{10}$ & 0.82$_{\pm 0.02}$ & \textbf{0.86$_{\pm 0.03}$} & 0.77$_{\pm 0.05}$ & 114.2$_{\pm 11.6}$ & 106.4$_{\pm 26.1}$ & 115.5$_{\pm 14.9}$ & .0023$_{\pm .0019}$ & .0055$_{\pm .0030}$ & .0059$_{\pm .0019}$ \\
 & $2^{13}$ & 1.31$_{\pm 0.03}$ & 1.33$_{\pm 0.02}$ & 1.26$_{\pm 0.04}$ & 43.8$_{\pm 15.2}$ & 39.3$_{\pm 15.3}$ & 55.0$_{\pm 18.5}$ & .0033$_{\pm .0018}$ & .0031$_{\pm .0022}$ & .0023$_{\pm .0005}$ \\
 & $2^{16}$ & 1.63$_{\pm 0.01}$ & 1.66$_{\pm 0.03}$ & 1.64$_{\pm 0.03}$ & 18.4$_{\pm 12.4}$ & 10.7$_{\pm 6.0}$ & 10.0$_{\pm 4.3}$ & .0011$_{\pm .0009}$ & .0063$_{\pm .0018}$ & .0016$_{\pm .0017}$ \\
\addlinespace[2pt]
Lotka--Volterra & $2^{10}$ & -2.59$_{\pm 0.01}$ & -2.57$_{\pm 0.03}$ & -2.60$_{\pm 0.01}$ & 77.1$_{\pm 4.1}$ & 83.7$_{\pm 15.9}$ & 83.1$_{\pm 24.2}$ & .0060$_{\pm .0012}$ & .0064$_{\pm .0013}$ & .0054$_{\pm .0006}$ \\
 & $2^{13}$ & -2.18$_{\pm 0.02}$ & -2.17$_{\pm 0.02}$ & -2.20$_{\pm 0.02}$ & 38.1$_{\pm 13.7}$ & 29.9$_{\pm 2.4}$ & 30.4$_{\pm 6.2}$ & .0065$_{\pm .0012}$ & .0071$_{\pm .0012}$ & .0065$_{\pm .0009}$ \\
 & $2^{16}$ & -1.98$_{\pm 0.02}$ & -1.99$_{\pm 0.02}$ & -2.00$_{\pm 0.01}$ & 18.3$_{\pm 16.1}$ & 23.0$_{\pm 15.4}$ & 18.3$_{\pm 11.7}$ & .0053$_{\pm .0012}$ & .0068$_{\pm .0011}$ & .0050$_{\pm .0016}$ \\
\bottomrule
\end{tabular}}
\end{table}

 \subsection{Imaging inverse problems} \label{sec:appendix-ip}

\paragraph{Problems} We selected three inverse problems introduced in \textsc{InverseBench} \citep{zheng2025inversebench}. They all have the form $y = \mathcal{A}(\theta) + \varepsilon$, where $y$ are measurements, $A$ is a forward model, $\theta$ is the signal we seek to reconstruct and $\varepsilon$ additive noise. 
\begin{itemize}
    \item \emph{Navier--Stokes (NS).} $\theta$ is the initial vorticity of a 2D incompressible flow on $(0,2\pi)^2$ with periodic boundary conditions ($128\times128$ pixels, Reynolds number 200, forcing $f(x)=-4\cos(4x_2)$), obtained by evolving a draw from $\mathcal N\big(0,(-\Delta+9I)^{-4}\big)$ for five time units. The forward operator is $\mathcal A=P\circ L$, where the solution operator $L$, computed with a black-box pseudo-spectral solver, maps $\theta$ to the vorticity at time $T=1$ and is nonlinear. The subsampling operator $P$ keeps one grid point out of two in each direction, giving $64\times64$ noise-free observations.
    
    \item \emph{Linear inverse scattering (IS).} $\theta$ is a $128\times128$ permittivity contrast map of HL60 cell nuclei, generated with the simulator of \citet{wiesner2019cytopacq}. Under the first Born approximation, $\mathcal A=A$ is linear, with $20$ transmitters and $360$ receivers on a circle of radius $1.6$\,m, and noise level $\sigma_y=10^{-4}$, resampled at every training batch.
    
    \item \emph{Compressed sensing MRI (MRI).} $\theta$ is a complex $320\times320$ knee image from fastMRI  \citep{zbontar2018fastmri}, a coil-combined minimum-variance unbiased estimate \citep{jalal2021robust}, stored as two real channels. Since these images have no coil sensitivity maps, we use the single-coil operator $\mathcal A=PF$, with $F$ the orthonormal 2D Fourier transform and $P$ a $\times4$ subsampling mask that keeps $80$ of the $320$ vertical k-space lines ($25$ central lines and $55$ random outer lines, redrawn for every example). The measurements are noise-free.

\end{itemize}

We did not select the \emph{Black hole imaging} problem because no training set of images is publicly available, and we did not select \emph{Full waveform inversion}, because it did not fit the shared setup of conditioning architecture with fixed image-shape. Our method is nonetheless applicable to these problems as well.

\paragraph{Noise-free measurements} On MRI and NS, measurements are noise-free, so the posterior is supported on the set $\{\theta:\mathcal A(\theta)=y\}$ and has no density on $\mathbb R^{d_\theta}$, which \cref{ass:A} requires. For the linear MRI operator, writing $\theta=A^+y+z$ with $z=(I-A^+A)\theta$, the range component $A^+y$ is fixed by the measurement, and the posterior is a distribution over the null-space component $z$, which has a density on the null space of $A$. Restricted to this subspace, the null-space distance $d(\theta,\theta')=\|(I-A^+A)(\theta-\theta')\|$ is a norm, so \cref{thm:proper} applies to the posterior of $z$ given $y$, with $\Theta$ replaced by the null space. This is why \tcvm uses this distance on MRI. On NS, the solution set of the nonlinear operator has no such linear parametrization, and \cref{thm:proper} does not strictly apply. There, the experiment tests the loss beyond its assumptions, and the Mira score, which requires no density, still measures coverage. On IS, the measurement noise gives the posterior a density, as \cref{ass:A} requires.

\paragraph{Diffusion models} 

Diffusion models \citep{sohl2015deep, ho2020denoising} map the target distribution $p(x)$ to white Gaussian noise by progressively perturbing the data $x$ through a diffusion process. Following \citet{song2021scorebased}, we express the continuous diffusion process as a stochastic differential equation (SDE):
\[
dx_t = f(x_t, t) dt + g(t)dw_t,
\]
where $x_t\in\mathbb{R}^N$ is the perturbed data at time $t$, $f(x_t, t)\in\mathbb{R}^N$ is the drift coefficient, $g(t)\in\mathbb{R}_+$ is the diffusion coefficient, and $w_t\in\mathbb{R}^N$ is a standard Wiener process. This SDE admits a reverse process, mapping noise to the target distribution:

\[
dx_t = \big[f(x_t, t) - \frac{1+\eta^2}{2} g(t)^2 \nabla_{x_t}\log p_t(x_t) \big]dt + \eta g(t) d\bar w_t,
\]
where $\bar w$ is a standard Wiener process when time flows backward. Among the different kinds of diffusion processes, we used the Variance Exploding schedule, which parametrizes the noising kernel $x_t\sim\mathcal{N}(\alpha_t x, \sigma_t^2 \mathbb{I}_N)$, with $\alpha_t=1$, $\log \sigma_t=(1-t)\log\sigma_{\min} + t \log \sigma_{\max}$, with $\sigma_{\min}=10^{-3}$ and $\sigma_{\max}=10^2$. We also adopt the preconditioning strategy of \citet{karras2022elucidating}, which parametrizes a denoising network $d_\phi$ instead  of the score function.

Our target distribution is the conditional distribution of $x=\theta\mid y$, where $y$ are measurements and $\theta$ is the signal we aim to recover. The measurement $y$ enters as a fixed image, concatenated channel-wise to the noisy input of the network. For both MRI and IS, we compute the adjoint solution to transform measurement into images $A^\top y$. NS observations are already $64\times64$ images, the resolution of the signal after $2\times2$ patching, so they are concatenated to the patched input without resampling. NS has no linear adjoint, so the switch rules use the nearest-neighbor upsampled observation as the point estimate $b(y)$. We describe all our choices for the network architectures and training strategies in Table~\ref{tab:diffusion-arch}. All fine-tuning runs share the calibration batch (16 pairs per step on NS, 8 otherwise), the sampler depth ($T=20$), the optimizer, the number of steps, and the sampler chains that carry gradients ($M_{\mathrm{grad}}=2$ per pair on NS and MRI, $8$ on inverse scattering). $\lambda$ is tuned once on Navier--Stokes and reused for IS and MRI ($\lambda=10$ for \tcvm and $\lambda=0.003$ for ES and pCRPS).

\begin{figure}[h]
    \centering
    Linear inverse scattering\\
    \includegraphics[width=0.8\linewidth]{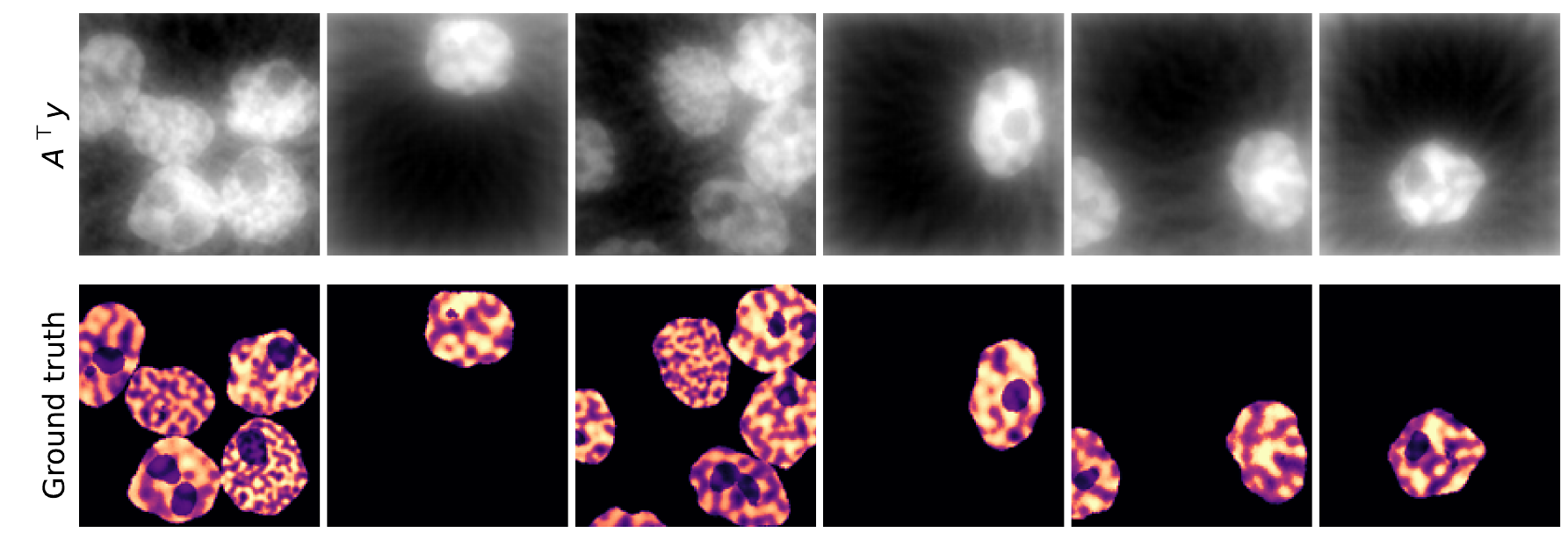}\\
    Navier--Stokes\\
    \includegraphics[width=0.8\linewidth]{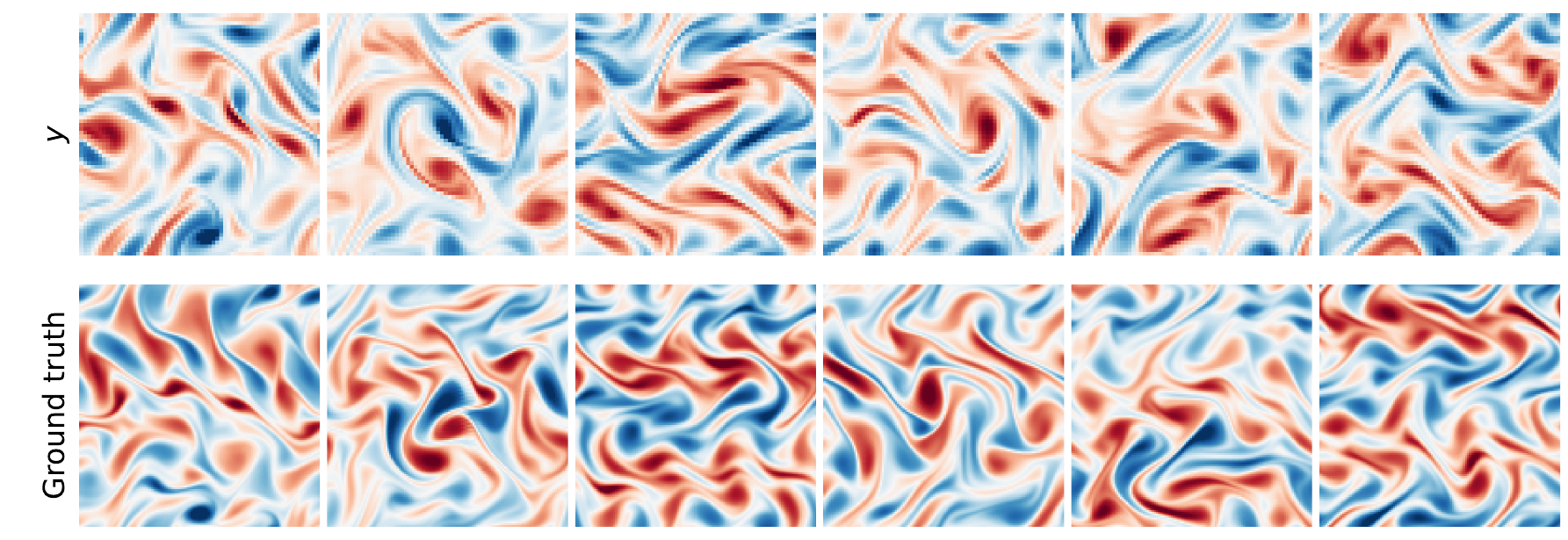}
    \\
    Compressed sensing MRI\\
    \includegraphics[width=0.8\linewidth]{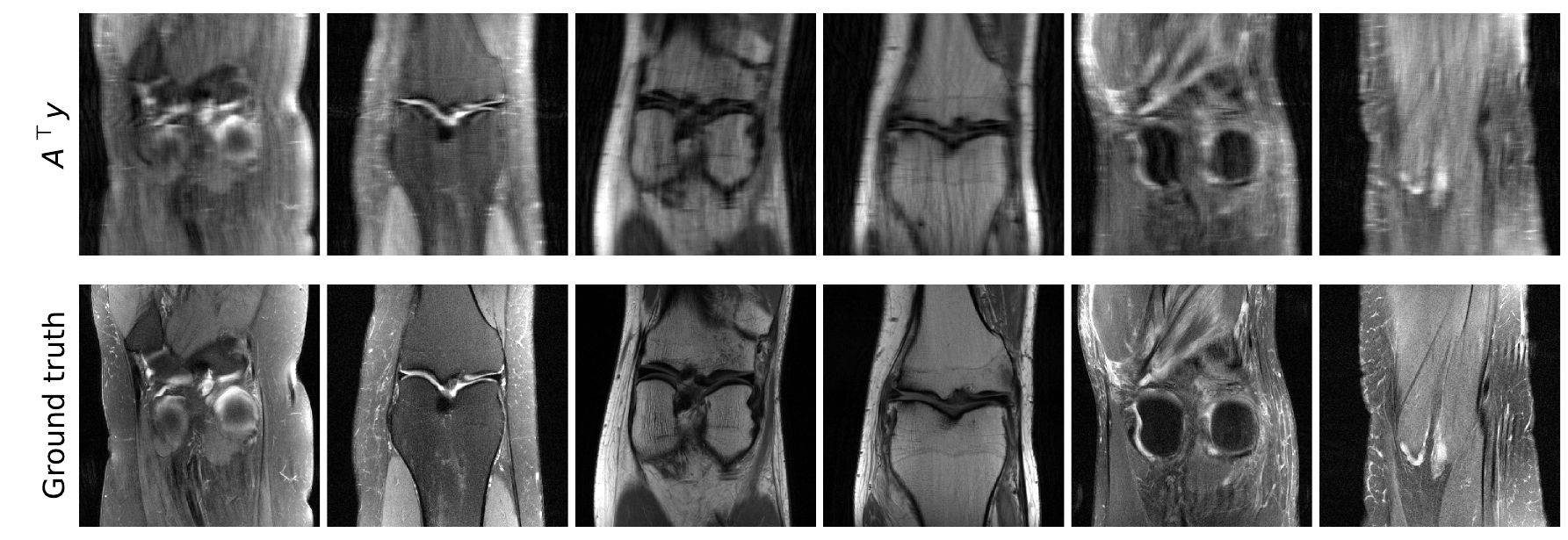}
    \caption{Examples of observations and target ground truth signals for the three inverse problems.}
    \label{fig:img}
\end{figure}

On one NVIDIA GH200, a fine-tuning step takes $1.9$\,s with \tcvm against $0.7$\,s with the energy score or pCRPS on NS, $1.0$\,s against $0.7$\,s on IS, and $3.7$\,s against $2.6$\,s on MRI, a $1.4$ to $2.7\times$ increase in training time with respect to the ES and pCRPS, and an $18$ to $33\times$ increase with respect to DSM alone. Figure \ref{fig:img} shows example observations and target ground truth signals for the three inverse problems.

The fastMRI data are used under the NYU fastMRI Dataset Sharing Agreement. All other code we use is available under open-source licenses (Apache 2.0, BSD 3-Clause or MIT), and the remaining datasets are publicly available.

\begin{table}[ht]
\centering
\caption{Architecture and training of the conditional diffusion posteriors. Fine-tuning starts from the converged networks from pretraining. All four fine-tuning objectives share every setting except the calibration weight $\lambda$.}
\label{tab:diffusion-arch}
\small
\begin{tabular}{lccc}
 & \textbf{Navier--Stokes} & \textbf{Inverse scattering} & \textbf{MRI} \\
\midrule
\multicolumn{4}{l}{\textit{Data}} \\
Signal $x$ shape                & $(128, 128, 1)$ & $(128, 128, 1)$ & $(320, 320, 2)$ \\
Conditioning                    & $y$, $(64, 64, 1)$ & $A^\top y$, $(128, 128, 1)$ & $[A^\top y,\ \text{mask}]$, $(320, 320, 3)$ \\
Train / validation images       & 19{,}456 / 1{,}024 & 8{,}500 / 1{,}500 & 22{,}964 / 2{,}048 \\
\midrule
\multicolumn{4}{l}{\textit{Architecture}} \\
Architecture                    & U-Net & U-Net & U-Net \\
Patch size                      & 2 & 2 & 4 \\
Input shape                     & $(64, 64, 5)$ & $(64, 64, 8)$ & $(80, 80, 80)$ \\
Residual blocks per level       & $(2, 2, 2, 2)$ & $(2, 2, 2, 2)$ & $(3, 3, 3, 3, 3)$ \\
Channels per level              & $(64, 128, 256, 256)$ & $(64, 128, 256, 256)$ & $(128, 256, 384, 512, 512)$ \\
Kernel size                     & 3 & 3 & 3 \\
Activation                      & SiLU & SiLU & SiLU \\
Normalization                   & LayerNorm & LayerNorm & LayerNorm \\
Noise embedding                 & 256 & 256 & 256 \\
Dropout                         & 0.1 & 0.1 & 0.1 \\
Parameters                      & 16.9\,M & 16.9\,M & 106.9\,M \\
\midrule
\multicolumn{4}{l}{\textit{Diffusion}} \\
Noise schedule                  & \multicolumn{3}{c}{VE, $\sigma \in [10^{-3}, 10^{2}]$, EDM preconditioning} \\
Training time distribution      & \multicolumn{3}{c}{$t \sim \mathrm{Beta}(3, 3)$} \\
Sampler                         & \multicolumn{3}{c}{DDIM ($\eta = 0$), 20 steps} \\
\midrule
\multicolumn{4}{l}{\textit{Pretraining (DSM)}} \\
Optimizer                       & Adam & Adam & Adam \\
Weight decay                    & 0.0 & 0.0 & 0.0 \\
Learning rate                   & $10^{-4}$ & $10^{-4}$ & $10^{-4}$ \\
Gradient norm clipping          & 1.0 & 1.0 & 1.0 \\
EMA decay                       & 0.999 & 0.999 & 0.999 \\
Augmentation                    & none & none & h-flip, roll $\pm 4$\,px \\
Batch size                      & 256 & 512 & 256 \\
Epochs (2 passes each)          & 560 & 580 & 204 \\
Optimizer steps                 & 85{,}120 & 18{,}560 & 36{,}312 \\
Early-stopping patience         & 10 & 10 & 20 \\
\midrule
\multicolumn{4}{l}{\textit{Fine-tuning}} \\
Optimizer                       & AdamW & AdamW & AdamW \\
Weight decay                    & 0.0 & 0.0 & 0.0 \\
Learning rate                   & $2 \times 10^{-5}$ & $2 \times 10^{-5}$ & $2 \times 10^{-5}$ \\
Gradient norm clipping          & 1.0 & 1.0 & 1.0 \\
EMA decay                       & 0.999 & 0.999 & 0.999 \\
DSM batch size                  & 64 & 64 & 64 \\
Optimizer steps                 & 24{,}000 & 24{,}000 & 24{,}000 \\
Calibration weight $\lambda$, ES and pCRPS & $0.003$ & $0.003$ & $0.003$ \\
Calibration weight $\lambda$, \tcvm        & $10$    & $10$    & $10$    \\
Calibration batch $N$           & 16 & 8 & 8 \\
Posterior samples $M$ (grad.\ + value) & $2 + 32$ & $8 + 16$ & $2 + 16$ \\
References $R$ (mixed)          & 128 & 64 & 64 \\
Temperature $\tau$              & \multicolumn{3}{c}{cosine, $0.1 \to 0.01 \times$ median pairwise posterior distance} \\
TARP distance                   & pixel, $2\times$ avg-pooled & pixel & null space of $A$ \\
Seeds                           & 3 & 3 & 3 \\
\bottomrule
\end{tabular}
\end{table}

\begin{table}[h]
\centering\footnotesize\setlength{\tabcolsep}{5pt}
    \caption{Quality metrics before and after calibration. Metrics on 64 held-out measurements with 32 posterior draws each  ($T=20$). PSNR, SSIM and relative $L_2$ of the posterior mean (the MMSE reconstruction) and PSNR of a single posterior draw, each against the ground truth. Spread is the standard deviation of the 32 draws at each pixel, averaged over pixels, in physical units.}
    \label{table:posterior_mean_quality}\begin{tabular}{llrrrrrr}
 & & \multicolumn{3}{c}{Posterior mean (MMSE)} & Draw & & \\
\cmidrule(lr){3-5}\cmidrule(lr){6-6}
Problem & Stage & PSNR\,$\uparrow$ & SSIM\,$\uparrow$ & rel.\,$L_2$\,$\downarrow$
        & PSNR & spread & DSM\,$\downarrow$ \\
\midrule
\multirow{5}{*}{NS}
  & DSM & 27.83 & 0.7851 & 0.2368 & 24.58 & 1.1169 & 0.0595 \\
  & $+$\,$\varnothing$ & 28.22\,$\pm$\,0.02 & 0.7969\,$\pm$\,0.0004 & 0.2280\,$\pm$\,0.0005 & 25.01\,$\pm$\,0.01 & 1.0606\,$\pm$\,0.0025 & 0.0594\,$\pm$\,0.0000 \\
  & $+$\,ES & 29.26\,$\pm$\,0.03 & 0.8246\,$\pm$\,0.0009 & 0.2061\,$\pm$\,0.0008 & 25.81\,$\pm$\,0.04 & 0.9847\,$\pm$\,0.0032 & 0.0590\,$\pm$\,0.0000 \\
  & $+$\,pCRPS & 29.23\,$\pm$\,0.04 & 0.8257\,$\pm$\,0.0010 & 0.2067\,$\pm$\,0.0008 & 25.79\,$\pm$\,0.04 & 0.9843\,$\pm$\,0.0037 & 0.0590\,$\pm$\,0.0000 \\
  & $+$\,TCvM & 27.88\,$\pm$\,0.05 & 0.7874\,$\pm$\,0.0011 & 0.2354\,$\pm$\,0.0013 & 24.22\,$\pm$\,0.05 & 1.2178\,$\pm$\,0.0101 & 0.0597\,$\pm$\,0.0000 \\
\midrule
\multirow{5}{*}{IS}
  & DSM & 29.62 & 0.9221 & 0.0906 & 27.93 & 0.0142 & 0.1159 \\
  & $+$\,$\varnothing$ & 29.79\,$\pm$\,0.01 & 0.9229\,$\pm$\,0.0002 & 0.0898\,$\pm$\,0.0000 & 28.23\,$\pm$\,0.01 & 0.0131\,$\pm$\,0.0000 & 0.1182\,$\pm$\,0.0001 \\
  & $+$\,ES & 29.72\,$\pm$\,0.02 & 0.9217\,$\pm$\,0.0003 & 0.0908\,$\pm$\,0.0002 & 28.21\,$\pm$\,0.01 & 0.0128\,$\pm$\,0.0000 & 0.1190\,$\pm$\,0.0001 \\
  & $+$\,pCRPS & 29.71\,$\pm$\,0.02 & 0.9216\,$\pm$\,0.0004 & 0.0909\,$\pm$\,0.0003 & 28.18\,$\pm$\,0.01 & 0.0129\,$\pm$\,0.0000 & 0.1194\,$\pm$\,0.0002 \\
  & $+$\,TCvM & 29.64\,$\pm$\,0.04 & 0.9224\,$\pm$\,0.0001 & 0.0905\,$\pm$\,0.0001 & 27.88\,$\pm$\,0.11 & 0.0179\,$\pm$\,0.0008 & 0.1186\,$\pm$\,0.0000 \\
\midrule
\multirow{5}{*}{MRI}
  & DSM & 37.11 & 0.8825 & 0.1232 & 35.10 & 0.0285 & 0.1916 \\
  & $+$\,$\varnothing$ & 37.47\,$\pm$\,0.00 & 0.8872\,$\pm$\,0.0000 & 0.1186\,$\pm$\,0.0000 & 35.41\,$\pm$\,0.00 & 0.0280\,$\pm$\,0.0000 & 0.1901\,$\pm$\,0.0000 \\
  & $+$\,ES & 37.79\,$\pm$\,0.01 & 0.8902\,$\pm$\,0.0003 & 0.1150\,$\pm$\,0.0001 & 34.71\,$\pm$\,0.06 & 0.0320\,$\pm$\,0.0003 & 0.1902\,$\pm$\,0.0000 \\
  & $+$\,pCRPS & 37.75\,$\pm$\,0.07 & 0.8905\,$\pm$\,0.0007 & 0.1153\,$\pm$\,0.0007 & 34.93\,$\pm$\,0.34 & 0.0331\,$\pm$\,0.0030 & 0.1902\,$\pm$\,0.0000 \\
  & $+$\,TCvM & 36.68\,$\pm$\,0.07 & 0.8751\,$\pm$\,0.0017 
    & 0.1291\,$\pm$\,0.0010 & 33.49\,$\pm$\,0.37 & 0.0443\,$\pm$\,0.0054 & 0.1910\,$\pm$\,0.0001 \\
\end{tabular}
\label{tab:quality}
\end{table}

\end{document}